\documentclass[11pt,a4paper]{article}

\usepackage[utf8]{inputenc}
\usepackage[T1]{fontenc}
\usepackage[english]{babel}

\usepackage[margin=1in]{geometry}
\usepackage{setspace}
\usepackage[expansion=false]{microtype}
\usepackage{parskip}
\usepackage{amsmath,amssymb,amsthm}
\usepackage{mathtools}

\usepackage{textcomp}

\usepackage{graphicx}
\graphicspath{{./}}
\usepackage{float}
\usepackage{booktabs}
\usepackage{array}
\usepackage{tabularx}
\usepackage{caption}
\usepackage{tikz}
\usetikzlibrary{arrows.meta,positioning,shapes.geometric,shadows,decorations.pathmorphing}
\usepackage[hidelinks,breaklinks=true]{hyperref}
\usepackage{xurl}

\usepackage{titlesec}
\titlespacing*{\section}{0pt}{1.4ex plus .2ex}{0.8ex plus .2ex}
\titlespacing*{\subsection}{0pt}{1.0ex plus .2ex}{0.6ex plus .2ex}
\titlespacing*{\subsubsection}{0pt}{0.8ex plus .2ex}{0.4ex plus .2ex}

\usepackage{fancyvrb}
\fvset{fontsize=\footnotesize,frame=single,framesep=2mm}

\makeatletter
\g@addto@macro\@floatboxreset\centering
\makeatother

\title{Architecture-Aware Explanation Auditing for Industrial Visual Inspection\\
\large A Model--Explainer--Perturbation Protocol for Faithfulness Assessment}
\author{}
\date{}
\begin{document}
\maketitle
\thispagestyle{empty}

\textbf{Sibo Jia}\textsuperscript{1} ------ \href{mailto:sibo.jia@ucdconnect.ie}{sibo.jia@ucdconnect.ie} \textbf{Zihang Zhao}\textsuperscript{2} ------ \href{mailto:zihang_zhao@emails.bjut.edu.cn}{zihang\_zhao@emails.bjut.edu.cn}

\textbf{Kunrong Li}\textsuperscript{2} ------ \href{mailto:lkr1213@emails.bjut.edu.cn}{lkr1213@emails.bjut.edu.cn}

\textsuperscript{1} \emph{Dublin College, Beijing University of Technology, Beijing, P.R. China}

\textsuperscript{2} \emph{College of Computer Science, Beijing University of Technology, P.R. China}
\begin{abstract}
Industrial visual inspection systems increasingly rely on deep classifiers whose heatmap explanations may appear visually plausible while failing to identify the image regions that actually drive model decisions. This paper operationalizes an \textbf{architecture-aware explanation audit protocol} grounded in the \textbf{native-readout hypothesis}: under a specified perturbation protocol, the measured faithfulness of an explanation method is constrained by its structural distance from the model's native decision mechanism. On WM-811K wafer maps (9 classes, 172 k images) under a three-seed zero-fill perturbation protocol, ViT-Tiny + Attention Rollout attains Deletion AUC 0.211 against 0.432--0.525 for Swin-Tiny / ResNet18+CBAM / DenseNet121 + Grad-CAM (|Cohen's \emph{d}| > 1.1), despite lower classification accuracy. Swin-Tiny disentangles architecture family from readout structure: despite being a Transformer, its spatial
feature-map hierarchy makes it Grad-CAM compatible, showing that the operative factor is \textbf{readout structure} rather than architecture family. A model-agnostic control (RISE) compresses all families to Deletion AUC $\approx$ 0.1, indicating that the native-method gap arises from the explainer pathway; notably, RISE outperforms all native methods, so native readout is a compatibility principle rather than an optimality guarantee. We therefore do not claim a universal ranking of explanation methods. Instead, explanation faithfulness is treated as a protocol-conditioned property of the model--explainer--perturbation triple rather than an intrinsic property of a heatmap method alone. A blur-fill sensitivity analysis shows that the family ordering reverses under a different perturbation baseline, reinforcing that faithfulness rankings are joint properties of (model, explainer, perturbation operator) triples. An exploratory boundary-condition study on MVTec AD (pretrained
models) indicates that audit results are dataset/task dependent and identifies conditions requiring qualification. The protocol yields actionable guidance: explanation pathways should be co-designed with model architectures based on readout structure, and deployed heatmaps should be accompanied by quantitative faithfulness metrics, model-agnostic audit baselines, and explicit perturbation-protocol reporting.
\end{abstract}

\section{Introduction}

\subsection{Motivation}

Deep learning is now deployed in decision settings where the prediction alone is no longer a sufficient artefact --- medical diagnosis, autonomous driving, credit scoring, and industrial quality inspection all require an accompanying explanation that a downstream reviewer (clinician, engineer, auditor) can interrogate. A standard operating procedure has emerged in response: train the strongest available classifier, then attach a post-hoc explanation method and present its heatmap as a window into the model's reasoning. This pairing treats the explanation method as a \emph{neutral measurement instrument}, independent of the classifier it examines. Under that assumption, the practitioner's only concern is to pick a method that is computationally cheap and visually clear; the choice of Grad-CAM vs. Integrated Gradients vs. LIME vs. attention-based methods is reduced to a convenience
decision. The evidence assembled in this paper suggests that assumption is systematically, and measurably, wrong. The explanation method is not neutral with respect to the model it is applied to; its \emph{faithfulness} --- the degree to which its output tracks prediction-relevant evidence under a specified evaluation protocol --- varies by more than a factor of two depending on whether its mathematical machinery matches the model's internal decision mechanism.

This problem aligns with the Industry 5.0 transition toward human-centric, sustainable, and resilient industry [21], where AI systems are expected to support human validation rather than only maximize predictive efficiency. In manufacturing and industrial cyber--physical systems, XAI is increasingly viewed as a mechanism for improving trustworthiness and reliability by enabling human operators to understand and validate AI decisions [22]. In this context, explanation maps in visual inspection should be treated as audit artifacts: they must help engineers verify whether a classifier relies on meaningful defect evidence rather than spurious spatial priors.

To study this claim rigorously, a testbed is needed in which the model families under comparison differ sharply in architecture, the ground-truth label signal is well-defined, the dataset is small enough that multi-seed replication is computationally tractable, and the practical stakes of a misleading explanation are non-trivial. WM-811K semiconductor wafer maps [18] satisfy all four criteria. The benchmark comprises 811,457 wafer maps, of which 172,950 carry a human-assigned defect-pattern label spanning nine classes, and has become the standard testbed for wafer-map defect classification. Contemporary industrial XAI practice on this benchmark typically proceeds exactly as described above --- train a high-capacity model, attach a post-hoc explainer, display the heatmap --- with faithfulness left undefended or asserted qualitatively. WM-811K therefore serves in this paper as an
instructive \emph{case study} for the broader XAI methodology question, not as an end in itself: the architectural claims and the statistical protocol developed below are intended to be testable wherever a deep classifier is paired with a post-hoc explainer, regardless of the imaging domain.

\subsection{The native-readout hypothesis}

This paper develops and tests a mechanistic hypothesis, termed the \textbf{native-readout hypothesis}: under a documented perturbation protocol, the faithfulness of an explanation method is constrained by its structural distance from the native decision mechanism of the model it explains. Throughout this paper, "native readout" refers to an explanation method whose mathematical machinery directly reads internal operators that participate in the model's forward decision pathway (e.g., attention matrices), as opposed to a "post-hoc proxy" that reconstructs importance from external signals (e.g., gradients of activations). Note that "native" denotes operator-alignment rather than causal exactness or literal identity with the forward computation: Attention Rollout, for instance, averages over heads and recursively multiplies across layers, which is not what the model computes during inference, but it reads from the same attention
operators that route information in the forward pass. "Structural distance" is operationalised through two concrete proxies. The first is whether the explanation pathway \emph{reads directly} from the operators that route information inside the model (as Attention Rollout does from the Transformer's attention matrices) or \emph{approximates them from the outside} through gradients (as Grad-CAM does through convolutional activations). The second is the spatial granularity at which the explanation is produced --- the patch scale for a Vision Transformer versus the effective receptive field of a deep convolutional block.

Under this hypothesis, five predictions follow. First, explanations that \emph{read natively} should produce lower Deletion AUC and higher Insertion AUC than explanations that approximate the decision mechanism post-hoc. Second, the faithfulness ranking of families should be \emph{insensitive} to their classification ranking, because the two quantities measure structurally independent properties. Third, the faithfulness ranking should be stable across random seeds, because it reflects architectural structure rather than learned weight configuration. Fourth, a \emph{model-agnostic} explainer --- one that treats the model as a black box --- should greatly compress the faithfulness ranking across families, because it bypasses the native-readout pathway entirely; if the gap persists under a model-agnostic method, the effect is attributable to the architecture's representations rather than
the explainer. Fifth, a \emph{hierarchical transformer} --- one that uses self-attention but maintains spatial feature maps through a multi-stage pyramid --- should align with Grad-CAM rather than Attention Rollout, because its spatial hierarchy provides a direct gradient-based readout path despite the underlying computation being attention-based; this prediction disentangles architecture family (CNN vs Transformer) from readout structure (spatial hierarchy vs global attention). We refer to this as the native-readout hypothesis, but use the term in a diagnostic rather than normative sense. It is not a formal theory of explanation optimality; rather, it is a compatibility principle predicting when an architecture-specific explainer is likely to be relatively faithful under a specified perturbation protocol. It does not imply that native explanations are always more faithful than
model-agnostic perturbation methods such as RISE. In this paper, RISE is therefore used as an offline audit reference rather than as a competing native explanation pathway.

The remainder of the paper evaluates these five predictions on WM-811K and tests boundary conditions on MVTec AD.

\subsection{Contributions}

The contributions of this paper are fourfold.

\textbf{(i) Hypothesis and audit protocol.} The paper formulates the native-readout hypothesis --- a falsifiable structural claim linking explanation faithfulness to architecture--explainer compatibility --- and operationalizes it as a reusable audit protocol for industrial visual inspection, with actionable guidance for selecting model--explainer pairs. A statistically rigorous evaluation design (per-class disaggregation, pooled Cohen's \emph{d}, bootstrap CIs, three-seed replication) strengthens inference from a modest sample budget.

\textbf{(ii) Multi-family empirical audit on WM-811K.} A quantitative cross-family faithfulness audit of four model families --- attention-free CNN, attention-augmented CNN, hierarchical transformer, and global-attention transformer --- fills a gap left by the wafer-XAI literature that presents heatmaps qualitatively. Swin-Tiny serves as a controlled falsification test that disentangles architecture family from readout structure, showing that spatial hierarchy rather than the convolution/attention distinction is the operative factor for Grad-CAM compatibility.

\textbf{(iii) Controls and ablations.} A model-agnostic control (RISE) compresses the family-level faithfulness spread, indicating the gap arises from the explainer pathway. A ViT-specific 2$\times$2 ablation design decomposes the ViT advantage into readout directness and multi-layer path depth. A blur-fill perturbation-baseline sensitivity analysis shows that the family ordering is protocol-dependent, reinforcing that faithfulness rankings are joint properties of (model, explainer, perturbation operator) triples.

\textbf{(iv) Boundary-condition analysis.} An exploratory study on MVTec AD (pretrained models, 256$\times$256 RGB) probes boundary conditions rather than serving as a matched replication, showing that audit results are dataset/task/architecture dependent and identifying conditions under which the hypothesis requires qualification.

\section{Related Work}

\subsection{Wafer-map XAI}

Interest in XAI for wafer-map classification remains a hot topic in 2025--2026. Khatun et al. [13] present a 0.15 M-parameter CBAM-enhanced CNN that reaches 99.88 \% test accuracy on a balanced WM-811K subset and pairs predictions with Grad-CAM, Integrated Gradients [11], and occlusion sensitivity [12]. Lee et al. [14] combine CNN, Grad-CAM, LIME, and temperature scaling to report accuracy, calibration, and dual visual explanations simultaneously. Park and Kim [16] take the opposite route, replacing CNNs with a Fuzzy Inference System that is intrinsically interpretable and robust to label noise. Lee et al. [15] show that a CNN ensemble with Radon- and density-based features reaches 95.09 \% accuracy on WM-811K. These contributions, along with Pilli's [17] thesis on human-in-the-loop XAI for wafer inspection, share a structural limitation: each pairs \emph{a} classifier with \emph{an}
explainer and reports heatmaps qualitatively, without asking whether the explainer and classifier are mechanistically compatible.

\subsection{Faithfulness in the broader XAI literature}

Within the broader interpretability literature, the faithfulness--plausibility gap has been documented in NLP by Jain and Wallace [9], who showed that attention weights in recurrent models are often unfaithful to the underlying decision. Chefer et al. [10] addressed this gap for Vision Transformers by combining Layer-wise Relevance Propagation with attention gradients. Attention Rollout [6] and the Deletion/Insertion perturbation protocols [7, 8] provide the measurement tools used in this paper. What has not been done, to the authors' knowledge, is a unified test on an industrial imaging benchmark of whether the structural distance between an explanation method and its model's decision mechanism predicts the faithfulness the method achieves. This paper addresses that gap.

\subsection{Model-agnostic explanations and cross-dataset validation}

RISE (Randomized Input Sampling for Explanation) [7] generates importance maps by probing the model with randomly masked inputs and weighting masks by the model's output confidence. Because it treats the model as a pure black box, it provides a principled control for disentangling architecture-driven from explainer-driven faithfulness differences. MVTec AD [19] is a widely-used benchmark for industrial anomaly detection comprising 5,354 images across 15 object categories with pixel-level defect annotations. Its higher resolution (700--1024 px), natural RGB imagery, and ground-truth masks make it a complementary testbed to WM-811K's low-resolution grayscale wafer maps. DeiT [20] provides data-efficient Vision Transformer variants pretrained on ImageNet, enabling fair comparison with pretrained CNNs on natural-image tasks.

\subsection{Trustworthy AI in industrial visual inspection}

The transition from Industry 4.0 automation toward Industry 5.0 emphasizes human-centric, sustainable, and resilient production systems, where AI is expected to support human validation rather than only maximize predictive efficiency [21]. Recent surveys of XAI in manufacturing and industrial cyber--physical systems similarly emphasize that interpretability is important for trustworthiness, reliability, and human validation of AI decisions in industrial settings [22]. Despite this recognition, existing industrial inspection papers overwhelmingly present heatmaps qualitatively --- showing that a method highlights "the right region" --- without auditing whether the highlighted pixels actually drive the model's prediction. The present work addresses this gap by operationalizing a quantitative faithfulness audit that can be applied to any classifier--explainer pair in industrial visual
inspection.

\begin{figure}[htbp]
\centering
\includegraphics[width=0.55\linewidth]{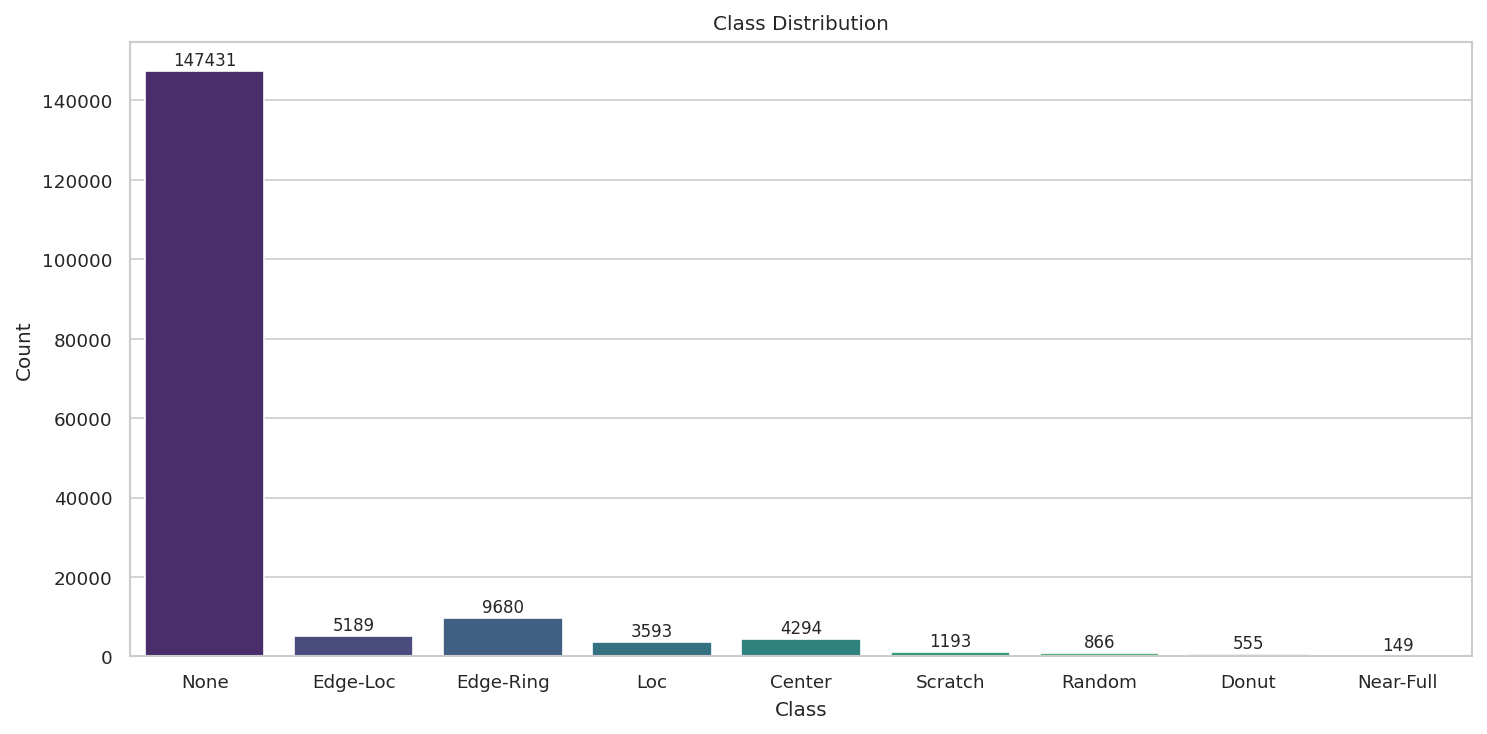}
\caption{\textbf{Figure 1.} WM-811K labelled class distribution}
\end{figure}

\begin{figure}[htbp]
\centering
\includegraphics[width=0.55\linewidth]{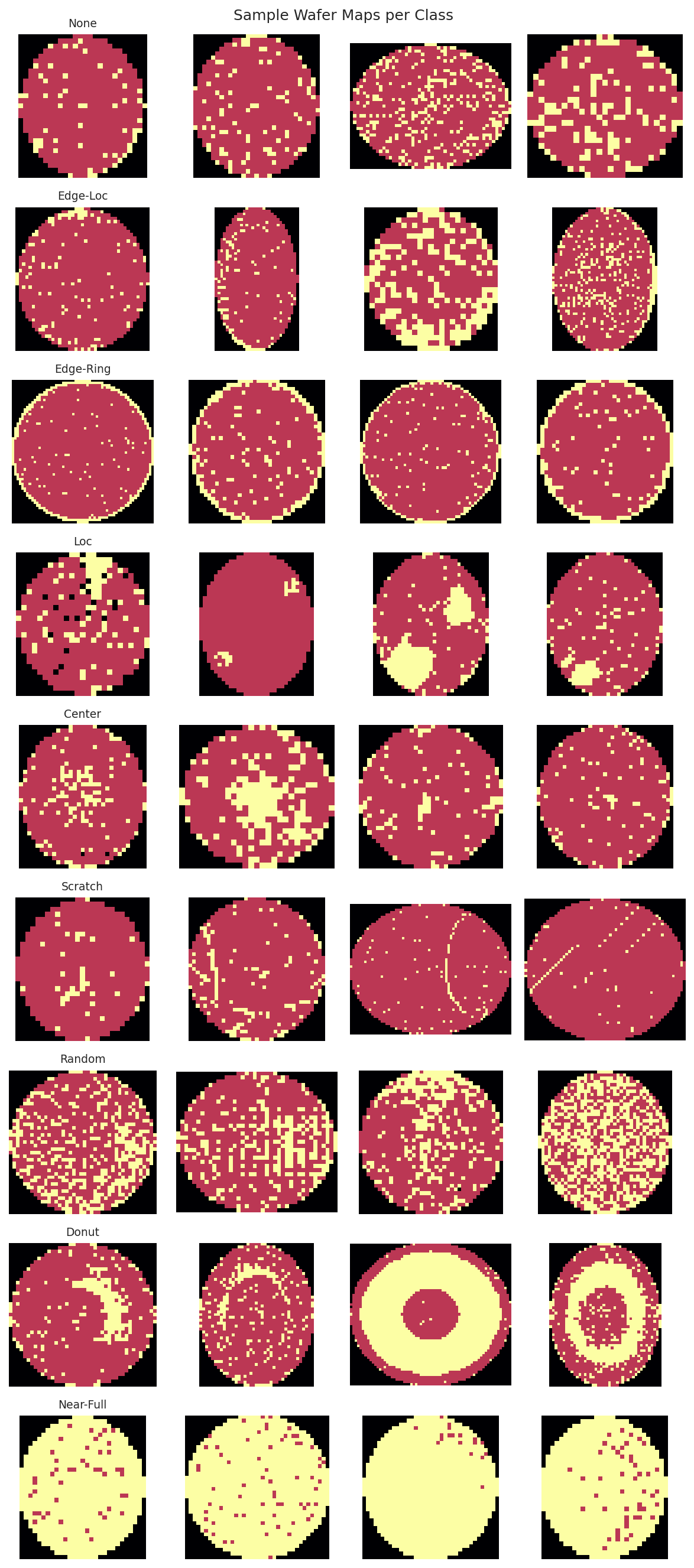}
\caption{\textbf{Figure 2.} Sample wafer maps (4 per class, 64$\times$64). Black = background, red = normal die, yellow = defect die}
\end{figure}

\section{Methods}

\subsection{Dataset}

The WM-811K benchmark [18] contains 811,457 wafer maps, of which 172,950 carry human-assigned defect-pattern labels spanning nine classes (None, Center, Donut, Edge-Loc, Edge-Ring, Loc, Random, Scratch, Near-Full). The dataset is strongly imbalanced, with approximately 85 \% of labelled maps belonging to the defect-free "None" class. Figures 1--2 show the class distribution and representative wafer maps. Because the majority class alone exceeds four-fifths of the labelled data, raw accuracy is an uninformative summary of model behaviour; this study therefore adopts balanced accuracy and macro-F1 as primary performance metrics, both of which treat each class equally and remain sensitive to performance on minority patterns.

The dataset is partitioned into training, validation, and test sets in a 70 / 15 / 15 ratio using a \textbf{lot-group split}. Under this scheme, all wafers originating from the same manufacturing lot are confined to a single partition, so that no lot is represented in more than one split. Lots are identified using the \texttt{lotName} field in the WM-811K metadata; samples without a valid lot identifier ($\approx$ 0.3 \%) are treated as singleton groups. This choice addresses the information-leakage risk that arises from the correlation between nearby wafers in the same lot: a simple random split would allow the model to exploit lot-specific defect patterns shared between training and test samples, inflating apparent test performance. Wafer maps are resized to 64$\times$64 pixels using nearest-neighbour interpolation, which preserves the discrete pixel semantics (background, pass, fail)
that bilinear or bicubic interpolation would smooth away. Training samples are augmented with random rotations, horizontal and vertical flips, and additive Gaussian noise ($\sigma$=0.05). Minority classes with fewer than 1,000 training samples receive a doubled augmentation probability; this mechanism partially offsets the class imbalance without synthesising new samples.

\subsection{Model architectures}

Four architecturally distinct families are compared, chosen to span the space of attention-free, attention-augmented, spatial-hierarchy transformer, and global-attention transformer designs. This four-point design isolates the role of both attention mechanism and spatial structure in the decision pathway --- from absent (DenseNet121), to auxiliary (ResNet18+CBAM), to constitutive-but-local (Swin-Tiny), to constitutive-and-global (ViT-Tiny) --- enabling a controlled test of how structural proximity between the explanation method and the native decision mechanism affects faithfulness. Crucially, the inclusion of Swin-Tiny disentangles architecture family (CNN vs Transformer) from readout structure (spatial hierarchy vs global attention): Swin uses self-attention but maintains hierarchical spatial feature maps, allowing a direct test of whether Grad-CAM compatibility depends on convolutions
or on spatial structure.

\subsubsection{ResNet18 + CBAM}

ResNet-18 [1] provides a strong convolutional baseline whose core building block implements a residual mapping,

\begin{equation}
\mathbf{y} = \mathcal{F}(\mathbf{x}, \{W_i\}) + \mathbf{x}
\tag{1}
\end{equation}

in which $\mathcal{F}$ is a stack of convolution, batch-normalisation, and ReLU layers. A Convolutional Block Attention Module (CBAM [2]) is appended after each of the four residual stages. CBAM refines an incoming feature map in two sequential steps. A channel-attention branch computes

\begin{equation}
M_c(\mathbf{F}) = \sigma\!\left(\text{MLP}(\text{AvgPool}(\mathbf{F})) + \text{MLP}(\text{MaxPool}(\mathbf{F}))\right)
\tag{2}
\end{equation}

followed by a spatial-attention branch,

\begin{equation}
M_s(\mathbf{F}') = \sigma\!\left(f^{7{\times}7}\!\left([\text{AvgPool}(\mathbf{F}');\, \text{MaxPool}(\mathbf{F}')]\right)\right)
\tag{3}
\end{equation}

yielding $\mathbf{F}'' = M_s(\mathbf{F}') \odot \mathbf{F}'$ where $\mathbf{F}' = M_c(\mathbf{F}) \odot \mathbf{F}$. Because both gating signals are explicit architectural components, this family exemplifies the "attention-by-design" approach to feature selection.

\subsubsection{DenseNet121}

DenseNet121 [3] uses dense inter-layer connectivity rather than explicit attention. Within a dense block, layer $\ell$ receives the concatenation of all preceding feature maps:

\begin{equation}
\mathbf{x}_\ell = H_\ell\!\left([\mathbf{x}_0,\, \ldots,\, \mathbf{x}_{\ell-1}]\right)
\tag{4}
\end{equation}

where $H_\ell$ denotes a batch-normalisation, ReLU, and 3$\times$3 convolution sequence. The architecture contains no explicit attention mechanism and therefore functions as a clean black-box convolutional baseline: any interpretability signal must be recovered post-hoc from gradients or perturbations.

\subsubsection{ViT-Tiny}

The Vision Transformer [4] processes the input as a sequence of non-overlapping patches, each linearly projected to a fixed embedding and concatenated with a learnable [CLS] token:

\begin{equation}
\mathbf{z}_0 = [\mathbf{x}_{\text{cls}};\; \mathbf{x}_1^p E;\; \ldots;\; \mathbf{x}_N^p E] + \mathbf{E}_{\text{pos}}, \quad N = HW / P^2
\tag{5}
\end{equation}

Each encoder layer applies multi-head self-attention,

\begin{equation}
\text{Attn}(Q, K, V) = \text{softmax}\!\left(\frac{QK^\top}{\sqrt{d_k}}\right)V
\tag{6}
\end{equation}

followed by a feed-forward network and residual connections. The configuration used here is patch size $P{=}4$ (yielding $N{=}256$ tokens for 64$\times$64 input), embedding dimension 192, depth 12, and 3 attention heads. ViT-Tiny possesses no convolutional inductive bias; all spatial relationships between patches emerge from the learned attention weights, making it the canonical "attention-first" architecture for the comparison.

\subsubsection{Swin-Tiny}

The Swin Transformer [23] bridges the CNN--ViT divide by combining self-attention with a hierarchical spatial structure. Unlike ViT, which applies global attention across all patches from layer 1, Swin restricts attention to local windows and introduces cross-window connections via shifted window partitioning:

\begin{equation}
\text{W-MSA}(\mathbf{z}^{\ell}) = \text{Concat}\left(\text{Attn}(Q_w, K_w, V_w)\right)_{w=1}^{M^2}
\tag{7}
\end{equation}

where each window $w$ contains $M{\times}M$ tokens (here $M{=}4$ for WM-811K at 64$\times$64, $M{=}7$ for MVTec at 256$\times$256). The architecture comprises four stages with progressive 2$\times$ spatial downsampling via patch merging, producing feature maps at resolutions $\frac{H}{4} {\times}\frac{W}{4}$, $\frac{H}{8}{\times}\frac{W}{8}$, $\frac{H}{16}{\times}\frac{W}{16}$, and $\frac{H}{32}{\times} \frac{W}{32}$. This hierarchical design yields spatial feature maps at each stage --- structurally analogous to ResNet's layer1--4 or DenseNet's denseblock1--4 --- despite the underlying computation being entirely attention-based.

This architectural choice has a direct consequence for explainability: Swin's final stage produces a spatial feature map (not a sequence of global tokens), making it a valid target for Grad-CAM --- though the final-stage resolution is coarse (2$\times$2 for 64$\times$64 input on WM-811K), so the expected benefit is structural compatibility rather than fine spatial precision. Swin-Tiny therefore occupies a diagnostic position in the experimental design: it is a Transformer (self-attention is the sole computational primitive) but its readout structure is spatial (hierarchical feature maps with progressive downsampling). If Grad-CAM compatibility depends on architecture family, Swin should behave like ViT; if it depends on readout structure, Swin should behave like the CNNs. This controlled contrast is the basis for Prediction 5 (\S{}1.2).

\begin{table}[htbp]
\centering
\small
\caption{\textbf{Table 1.} Model family summary}
\begin{tabular}{llrr}
\toprule
Family & Description & Parameters & Epochs (primary seed) \\
\midrule
ResNet18+CBAM & CNN + engineered attention & 11,219,793 & 31 \\
DenseNet121 & Pure CNN & 6,956,809 & 28 \\
ViT-Tiny & Pure Transformer & 5,393,289 & 82 \\
Swin-Tiny & Hierarchical Transformer & 27,504,723 & 90 \\
\bottomrule
\end{tabular}
\end{table}

\subsection{Interpretability methods}

Each family is paired with the most natural explanation method for its architecture, and all methods are evaluated under the same faithfulness protocol (\S{}3.4). The pairing is deliberate: Grad-CAM reads the final spatial feature maps that CNNs and hierarchical transformers use for spatial encoding, while Attention Rollout reads the attention matrices that constitute the global Transformer's native information-routing mechanism. A model-agnostic control (RISE) is additionally applied to all families to disentangle architecture-driven from explainer-driven faithfulness differences. This aims to maximise the faithfulness ceiling for each family and reduces the risk that observed differences are driven merely by a suboptimal method choice.

\subsubsection{Grad-CAM (for CNNs and Swin)}

Grad-CAM [5] produces a class-discriminative heatmap from the gradients flowing into a target convolutional layer. For target class $c$ and feature maps $A^k$ at the chosen layer:

\begin{equation}
\alpha_k^c = \frac{1}{Z} \sum_i \sum_j \frac{\partial y^c}{\partial A^k_{ij}}
\tag{8}
\end{equation}

\begin{equation}
L_{\text{Grad-CAM}}^c = \text{ReLU}\!\left(\sum_k \alpha_k^c A^k\right)
\tag{9}
\end{equation}

The ReLU retains only features with positive influence on class $c$; the result is upsampled to input resolution and min-max normalised to [0, 1]. Grad-CAM occupies a specific position in the native-readout framework: it is a \textbf{gradient-based proxy} for the CNN's decision mechanism rather than a direct readout of it. The CNN's native decision operators are its learned convolutional kernels and the spatial feature maps they produce; however, these feature maps are not individually interpretable as class-specific explanations --- hundreds of maps respond to different patterns simultaneously, and no built-in mechanism selects which spatial locations drove a particular class decision. Grad-CAM bridges this gap by back-propagating gradients of the target logit through those kernels and weighting the activations accordingly, but this reconstruction is inherently indirect. The target
layer choice partially controls the structural distance: \texttt{cbam4} is used for ResNet18+CBAM (the post-attention feature map consumed by the classifier), \texttt{features.denseblock4} for DenseNet121 (the last dense block before global pooling), and \texttt{layers[-1]} (the final transformer stage) for Swin-Tiny. All three choices minimise the distance subject to the constraint that no explicit routing signal is available to read. Notably, Swin's final stage is structurally analogous to a CNN's last convolutional block --- it produces a spatial feature map --- despite being computed entirely via windowed self-attention.

\subsubsection{Attention Rollout (for ViT)}

Attention Rollout [6] accumulates attention across all Transformer layers to estimate how much each input patch contributes to the final [CLS] token. At layer $\ell$, the raw attention matrix $A^{(\ell)} \in \mathbb{R}^{N \times N}$ is averaged across heads and combined with the identity matrix to account for the residual connection:

\begin{equation}
\hat{A}^{(\ell)} = 0.5 \cdot \bar{A}^{(\ell)} + 0.5 \cdot I
\tag{10}
\end{equation}

Rollout then recursively multiplies:

\begin{equation}
R^{(\ell)} = \hat{A}^{(\ell)} \cdot R^{(\ell-1)}, \quad R^{(0)} = I
\tag{11}
\end{equation}

The final heatmap is the [CLS] row of $R^{(L)}$, reshaped to the spatial grid and normalised. Attention Rollout occupies the opposite position in the native-readout framework: it is a \textbf{closer forward-pass readout} of the Transformer's information-routing structure than gradient-based post-hoc proxies. The Transformer's attention matrices are part of its native information-routing pathway, governing how information flows from patch tokens into the [CLS] token consumed by the classifier. Rollout accumulates those matrices through the residual stream (Eq. 10) and across all twelve layers via the recursive product (Eq. 11). No gradient is involved: the explanation is computed from the same forward pass that produced the prediction, and the quantity reported is a first-order property of the decision itself. Unlike Grad-CAM, Rollout is not explicitly class-discriminative --- it
estimates token contribution to the [CLS] pathway regardless of which class logit is highest. In this study it is evaluated against the predicted-class probability, so the faithfulness question is whether the [CLS] information-routing map identifies pixels the model uses for its own prediction. The more recent class-specific method of Chefer et al. [10] combines LRP with attention gradients; Rollout is retained here precisely \emph{because} it is the minimal-distance readout of the Transformer's routing matrices and therefore the cleanest instrument for testing the native-readout hypothesis.

Note that Attention Rollout is \emph{not} applied to Swin-Tiny. Swin lacks a global [CLS] token --- its classification head operates on the spatially-averaged output of the final stage --- and its windowed attention matrices do not define a global token-to-token routing path. This architectural incompatibility is itself evidence that Swin's readout structure is spatial rather than global, motivating the use of Grad-CAM for this family.

\subsubsection{RISE (model-agnostic control)}

RISE (Randomized Input Sampling for Explanation) [7] generates importance maps by probing the model with randomly masked inputs. $N$ binary masks are sampled at a coarse resolution $s{\times}s$ (default $s{=}8$), upsampled to input resolution via bilinear interpolation, and applied element-wise to the input image. The model's output confidence for the target class under each mask is recorded, and the final saliency map is the confidence-weighted average of all masks:

\begin{equation}
S(x) = \frac{1}{N} \sum_{i=1}^{N} f_c(x \odot M_i) \cdot M_i
\tag{12}
\end{equation}

where $f_c$ is the model's softmax probability for class $c$ and $M_i$ is the $i$-th upsampled mask. Because RISE treats the model as a pure black box --- no gradients, no internal activations, no architecture-specific hooks --- it provides a principled control for disentangling architecture-driven from explainer-driven faithfulness differences. If native methods outperform RISE, the advantage is attributable to the explainer's structural alignment with the model; if RISE matches or exceeds native methods, the native pathway provides no additional faithfulness benefit. In this study, $N{=}4000$ masks at resolution $8{\times}8$ are used for both WM-811K and MVTec AD evaluations.

\begin{table}[htbp]
\centering
\small
\caption{\textbf{Table 2.} Interpretability method assignment}
\begin{tabular}{llll}
\toprule
Family & Method & Target & Readout type \\
\midrule
ResNet18+CBAM & Grad-CAM (Eq. 8-9) & cbam4 & Post-hoc proxy (gradient) \\
DenseNet121 & Grad-CAM (Eq. 8-9) & features.denseblock4 & Post-hoc proxy (gradient) \\
Swin-Tiny & Grad-CAM (Eq. 8-9) & layers[-1] (final stage) & Post-hoc proxy (gradient) \\
ViT-Tiny & Attention Rollout (Eq. 10-11) & All 12 encoder layers & Native readout (forward) \\
All families & RISE (Eq. 12) & Black-box (N=4000 masks) & Model-agnostic \\
\bottomrule
\end{tabular}
\end{table}

Because each architecture is paired with its most natural explainer, the main comparison evaluates model--explainer pairs rather than architectures alone; disentangling architecture effects from explainer effects requires the ablations reported in \S{}S9 and \S{}S10 (discussed in \S{}5.1 and the Limitations section (\S{}5.7)).

\subsection{Training and evaluation protocol}

\subsubsection{Training}

All four families share a unified optimisation framework. Optimisation uses Adam or AdamW depending on the family's configuration, with a learning-rate schedule that is either cosine or reduce-on-plateau. Early stopping is triggered on validation balanced accuracy for the CNN families and validation macro-F1 for ViT-Tiny (following the family-specific configurations in Supplementary Table S1), which better reflects the imbalanced objective than raw loss. Because ViT-Tiny converges more slowly without convolutional inductive bias, it is allowed a larger maximum epoch budget; comparisons use the best validation checkpoint rather than the final epoch. Every training run writes its full configuration to \texttt{config.yaml} within the run directory, enabling exact reconstruction of any experiment from its stored artefacts. Complete per-family hyperparameters are provided in Supplementary
Table S1.

\subsubsection{Seeds and variance estimation}

To estimate run-to-run variance rather than report single-run point estimates, every family is trained with three random seeds, yielding twelve WM-811K runs in total. ResNet18+CBAM, DenseNet121, and ViT-Tiny use seeds 42, 123, and 456; Swin-Tiny uses seeds 7, 123, and 456 (seed 42 produced a degenerate initialisation for this family; see Supplementary \S{}S13). All metrics reported are aggregated as mean $\pm$ standard deviation across these seeds. Per-seed classification results are listed in Supplementary Table S3.

\subsubsection{Faithfulness metrics}

Explanation quality is evaluated using three complementary metrics, each computed identically for all families.

\textbf{Deletion AUC} [7] progressively zeros out pixels ranked highest by the heatmap and tracks the predicted-class probability:

\begin{equation}
\text{Del-AUC} = \int_0^1 P\bigl(c \mid x_{\setminus \text{top-}f}\bigr)\, df
\tag{13}
\end{equation}

A faithful explanation ranks decision-relevant pixels first; the probability drops quickly, yielding a \textbf{small} AUC.

\textbf{Insertion AUC} [7] starts from an all-zero image and progressively inserts top-ranked pixels:

\begin{equation}
\text{Ins-AUC} = \int_0^1 P\bigl(c \mid x_{\text{top-}f}\bigr)\, df
\tag{14}
\end{equation}

A faithful explanation recovers confidence with few pixels, yielding a \textbf{large} AUC.

\textbf{Stability} [8] measures heatmap consistency under semantics-preserving perturbations. For $K{=}5$ augmentations (rotation $\pm$15\textdegree{}, translation $\pm$3 px, Gaussian noise $\sigma$=0.02):

\begin{equation}
\text{Stability} = \frac{1}{K}\sum_{k=1}^K \frac{\langle h(x),\, h(x_k')\rangle}{\|h(x)\| \cdot \|h(x_k')\|}
\tag{15}
\end{equation}

computed over the wafer region only (pixels > 0). All perturbation curves are computed with respect to the model's predicted class, so the evaluation measures faithfulness to the model's own decision pathway rather than correctness relative to the labelled class. The zero-fill baseline is domain-motivated for WM-811K: wafer maps are sparse discrete die-state arrays rather than natural RGB images, so setting selected pixels to zero approximates removing local die-state evidence by replacing it with the background/non-wafer value. These perturbation metrics are treated as proxy measures of model--explainer faithfulness under the specified zero-fill protocol, rather than as pixel-level ground-truth causal explanations. A blur-fill sensitivity check (\S{}4.6) tests whether the family ordering is robust to this baseline choice.

Throughout this paper, unless otherwise stated, "faithfulness" denotes protocol-conditioned perturbation faithfulness: the extent to which a heatmap ranking identifies pixels whose removal or insertion changes the model's predicted-class probability under the specified perturbation operator. It is therefore a proxy for model--explainer alignment, not a direct measure of causal explanation truth.

\subsubsection{Sample size and statistical analysis}

The metrics are computed on 200 balanced-stratified test samples per seed (22 per class $\times$ 9 classes = 198 effective samples due to integer rounding; 594 per family across 3 seeds). Deletion and Insertion AUCs are computed over 20 perturbation steps spanning the full range from 0 to 100 \% of ranked pixels. For the supplementary top-k confidence- drop analysis (\S{}S12), the default threshold is 10 \% of pixels (410 of 4,096), which is fine enough to distinguish localised defects yet coarse enough to avoid sensitivity to single-pixel noise. Five augmentations per sample are the minimum needed to estimate cosine- similarity variance while keeping total evaluation time tractable on a single GPU. Full parameter settings are listed in Supplementary Table S2.

Beyond the aggregate mean $\pm$ std reporting conventional in prior wafer-XAI work, three complementary statistical analyses are used: (i) per-class disaggregation of the Deletion AUC to verify that findings are not driven by a favourable subset of classes; (ii) pooled Cohen's \emph{d} [effect size] quantifying the standardised magnitude of the family-level difference; and (iii) non-parametric bootstrap 95 \% confidence intervals (2,000 resamples) for the mean Deletion AUC of each family.

\subsection{Model-agnostic control: RISE}

To test the fourth prediction (\S{}1.2) --- that a model-agnostic explainer should greatly compress the faithfulness ranking --- RISE [7] (described in \S{}3.3.3) is applied uniformly to all four families on WM-811K. The same $N{=}4000$ masks at resolution $8{\times}8$ are used, and the resulting heatmaps are evaluated with the same Deletion/Insertion protocol as the native methods (\S{}3.4.3), enabling direct comparison. RISE hyperparameters ($N{=}4000$, mask resolution $8{\times}8$, probability $p{=}0.5$) are fixed identically across all families.

\subsection{Boundary-condition dataset: MVTec AD}

To probe boundary conditions beyond WM-811K, the same audit workflow is applied to MVTec AD [19] --- a benchmark for industrial anomaly detection comprising 5,354 images across 15 object categories (e.g., carpet, leather, metal nut, transistor) with pixel-level defect masks. This study is not a matched replication: it uses RGB natural images, ImageNet-pretrained backbones, a binary classification task, and a single evaluation seed.

\textbf{Task framing.} All categories are pooled into a binary classification task (good vs. defective: 4,096 / 1,258 samples). A 70/15/15 stratified random split yields 3,748 training, 803 validation, and 803 test samples.

\textbf{Preprocessing.} Images are resized to 256$\times$256 and normalised to ImageNet statistics. Data augmentation mirrors WM-811K (rotation, flip, Gaussian noise).

\textbf{Models.} The same four architectural families are used, all with ImageNet-pretrained backbones: ResNet18+CBAM (torchvision), DenseNet121 (torchvision), ViT-Tiny (DeiT-Tiny [20], timm), and Swin-Tiny [23] (timm). Pretraining is necessary because MVTec AD's natural RGB images and small sample size preclude effective from-scratch training for ViT (confirmed empirically: from-scratch ViT achieves only 55 \% balanced accuracy). All models are fine-tuned with AdamW, cosine LR schedule, and early stopping (patience 15). A single seed (42) is used for the boundary-condition study.

\textbf{Interpretability evaluation.} Native methods (Grad-CAM for CNNs and Swin, Attention Rollout for ViT) and RISE are applied to all families. Because MVTec provides pixel-level defect masks, IoU between the thresholded heatmap and the ground-truth mask is additionally computed as a spatial-alignment metric. Evaluation uses 200 defective test samples (seed 42).

\textbf{RISE mask-resolution ablation.} Because MVTec images are 4$\times$ larger than WM-811K (256 vs 64 px), the default $8{\times}8$ mask grid produces 32 px cells --- potentially too coarse for fine-grained defects. An ablation over mask resolutions $\{8, 16, 32\}$ tests whether RISE's performance is resolution-limited on this dataset.

All code, configurations, and evaluation scripts are publicly available (Supplementary \S{}S14).

\section{Results}

\subsection{Classification performance}

The classification performance of the four families, averaged over three random seeds, is reported in Table 3.

\begin{table}[htbp]
\centering
\small
\caption{\textbf{Table 3.} Classification performance (mean $\pm$ std across 3 seeds)}
\resizebox{\linewidth}{!}{%
\begin{tabular}{lrrrrr}
\toprule
Family & Accuracy & Balanced Acc & F1 Macro & F1 Weighted & MCC \\
\midrule
ResNet18+CBAM & 0.957 $\pm$ 0.004 & 0.901 $\pm$ 0.001 & 0.849 $\pm$ 0.005 & 0.960 $\pm$ 0.003 & 0.858 $\pm$ 0.011 \\
DenseNet121 & 0.962 $\pm$ 0.004 & 0.901 $\pm$ 0.007 & 0.853 $\pm$ 0.011 & 0.964 $\pm$ 0.003 & 0.872 $\pm$ 0.011 \\
ViT-Tiny & 0.928 $\pm$ 0.007 & 0.799 $\pm$ 0.014 & 0.747 $\pm$ 0.008 & 0.935 $\pm$ 0.005 & 0.774 $\pm$ 0.019 \\
Swin-Tiny & 0.953 $\pm$ 0.002 & 0.822 $\pm$ 0.015 & 0.792 $\pm$ 0.004 & 0.957 $\pm$ 0.002 & 0.838 $\pm$ 0.002 \\
\bottomrule
\end{tabular}
}
\end{table}

ViT-Tiny trails the CNN families by approximately 10 percentage points in macro-F1, while Swin-Tiny reduces this gap to roughly 6 points, consistent with the benefit of its hierarchical local-window structure. This gap is expected and well-documented [4, 23]: Vision Transformers lack the convolutional inductive bias (translation equivariance, local connectivity) that gives CNNs a strong spatial prior on small, low-resolution datasets. Without large-scale pretraining --- which is impractical on 64$\times$64 single-channel wafer maps --- transformers must learn all spatial relationships from scratch, requiring substantially more epochs (67--97 vs 28--36) and still converging to a weaker optimum. Swin-Tiny's local-window attention provides a partial inductive bias (locality within each window), placing it slightly above ViT-Tiny (82.2\% vs 79.9\% BAcc) but still below the CNNs. Crucially,
this classification disadvantage does not invalidate the interpretability comparison, but it does require the faithfulness results to be interpreted separately from predictive performance. The later commonly-correct subset and RISE controls therefore play an important role in reducing the risk that the observed faithfulness gap is merely a by-product of classification quality.

Per-class F1 is visualised as a radar chart (Figure 3), and per-family training and validation loss curves are provided in Supplementary Figure S1.

\begin{figure}[htbp]
\centering
\includegraphics[width=0.55\linewidth]{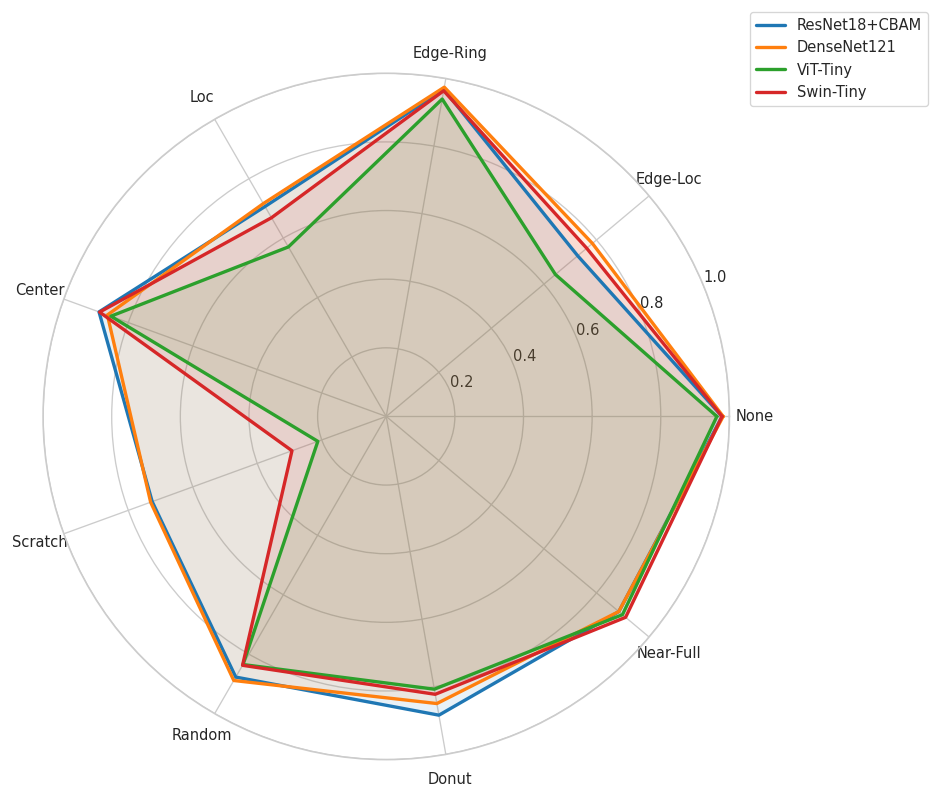}
\caption{\textbf{Figure 3.} Per-class F1 radar (mean across 3 seeds)}
\end{figure}

\subsection{Qualitative interpretability}

Before reporting the quantitative faithfulness metrics (\S{}4.3), this subsection examines the heatmaps qualitatively. Comparing interpretability methods across architecturally distinct families is methodologically delicate: Grad-CAM on a CNN and Attention Rollout on a Transformer derive from mathematically different objects --- gradient-weighted activations versus cumulative attention products --- so visual similarity between heatmaps is not directly commensurable. Under the native-readout hypothesis, however, this incommensurability is precisely the object of study. The evaluation therefore compares the \emph{output behaviour} of each method --- what drops when top-ranked pixels are removed, what grows when they are added back, how stable the ranking is under perturbation --- rather than the surface appearance of the heatmap. No attempt is made to recover a causally "true" explanation
for any individual prediction, as WM-811K provides no pixel-level ground truth. The claims supported here are of the form "family X's explanation tracks its predictions more faithfully than family Y's under deletion perturbation" --- measurable, reproducible, and hypothesis-relevant.

For each class, the highest-confidence correctly-classified test sample (using ResNet18+CBAM as reference) is shown in Figure 5, with Grad-CAM overlays for the two CNN families and Attention Rollout for ViT-Tiny. Before examining the full panel, Figures 4a--4c present a close-up of the \textbf{Center} class --- the pattern where the three methods diverge most visibly.

\begin{figure}[htbp]
\centering
\includegraphics[width=0.98\linewidth]{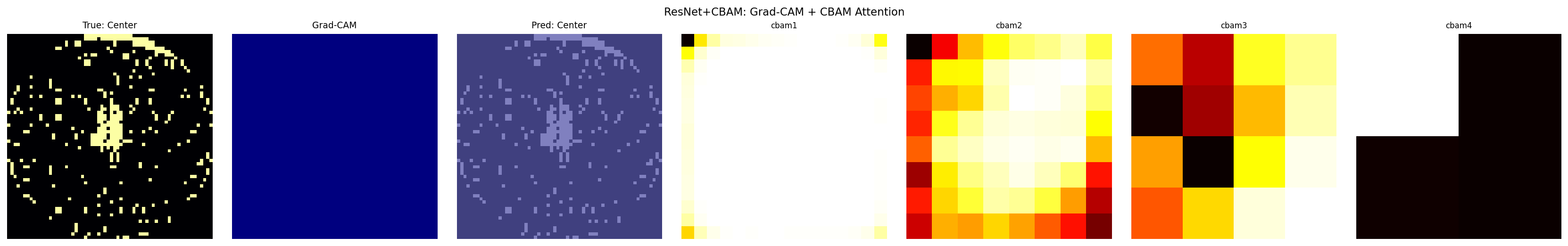}
\caption{\textbf{Figure 4a.} ResNet18+CBAM --- Center-class explanation. Left to right: original wafer, Grad-CAM heatmap, prediction overlay, and CBAM spatial attention at layers 1--4. The Grad-CAM map is nearly uniformly dark (degenerate), providing no spatial localisation of the central defect. The CBAM attention maps (cbam2--cbam3) show coarse quadrant-level patterns, illustrating a case where the post-hoc gradient signal provides little spatial localisation of the visible central defect pattern}
\end{figure}

\begin{figure}[htbp]
\centering
\includegraphics[width=0.98\linewidth]{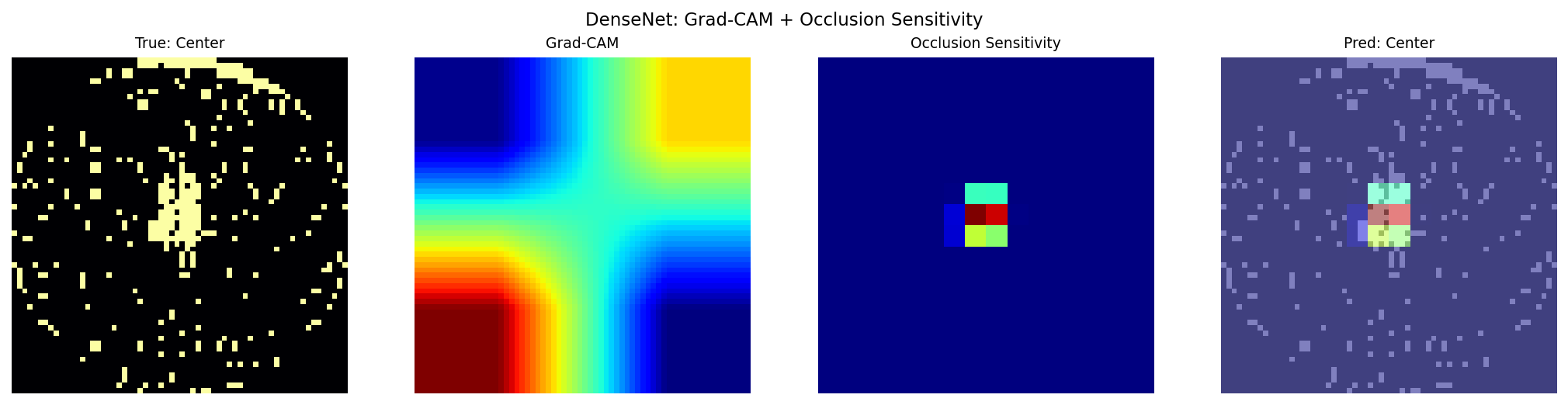}
\caption{\textbf{Figure 4b.} DenseNet121 --- Center-class explanation. Left to right: original wafer, Grad-CAM heatmap, occlusion sensitivity map, and prediction overlay. Grad-CAM places its hot zone in the lower-left corner --- far from the central defect cluster --- suggesting that the gradient proxy does not recover defect location for this sample. The occlusion sensitivity map (a model-agnostic reference) does localise the center, highlighting the gap between what the gradient path reports and where the model is actually sensitive}
\end{figure}

\begin{figure}[htbp]
\centering
\includegraphics[width=0.98\linewidth]{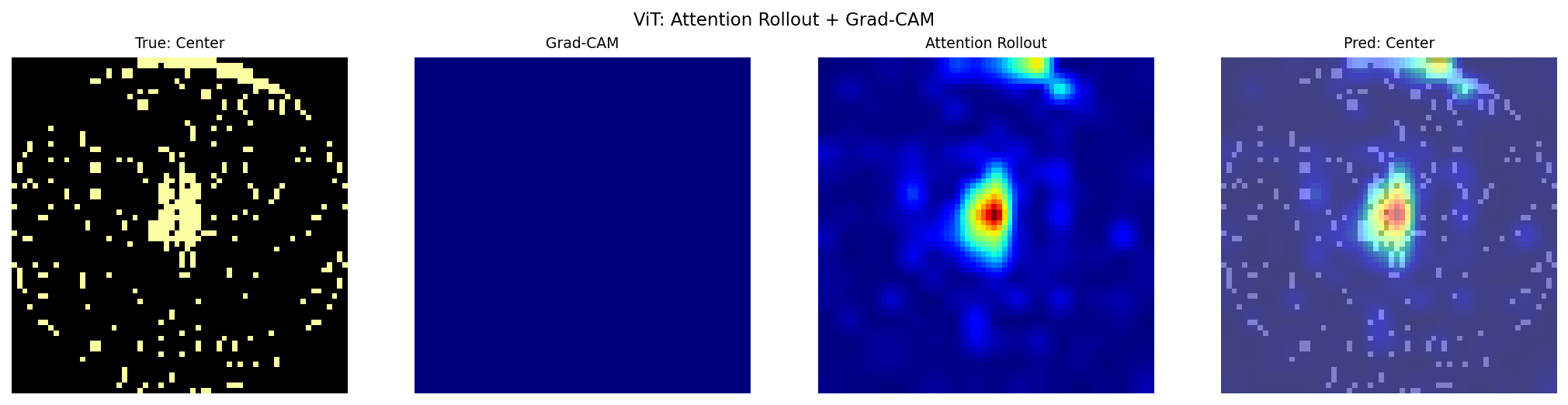}
\caption{\textbf{Figure 4c.} ViT-Tiny --- Center-class explanation. Left to right: original wafer, Grad-CAM (blank, as expected for a non-CNN architecture), Attention Rollout heatmap, and prediction overlay. Rollout concentrates activation on the central defect cluster, consistent with the native-readout hypothesis: the explanation is derived from the model's forward-pass attention pathway and spatially aligns with the defect pattern in this sample}
\end{figure}

\begin{figure}[htbp]
\centering
\includegraphics[width=0.60\linewidth]{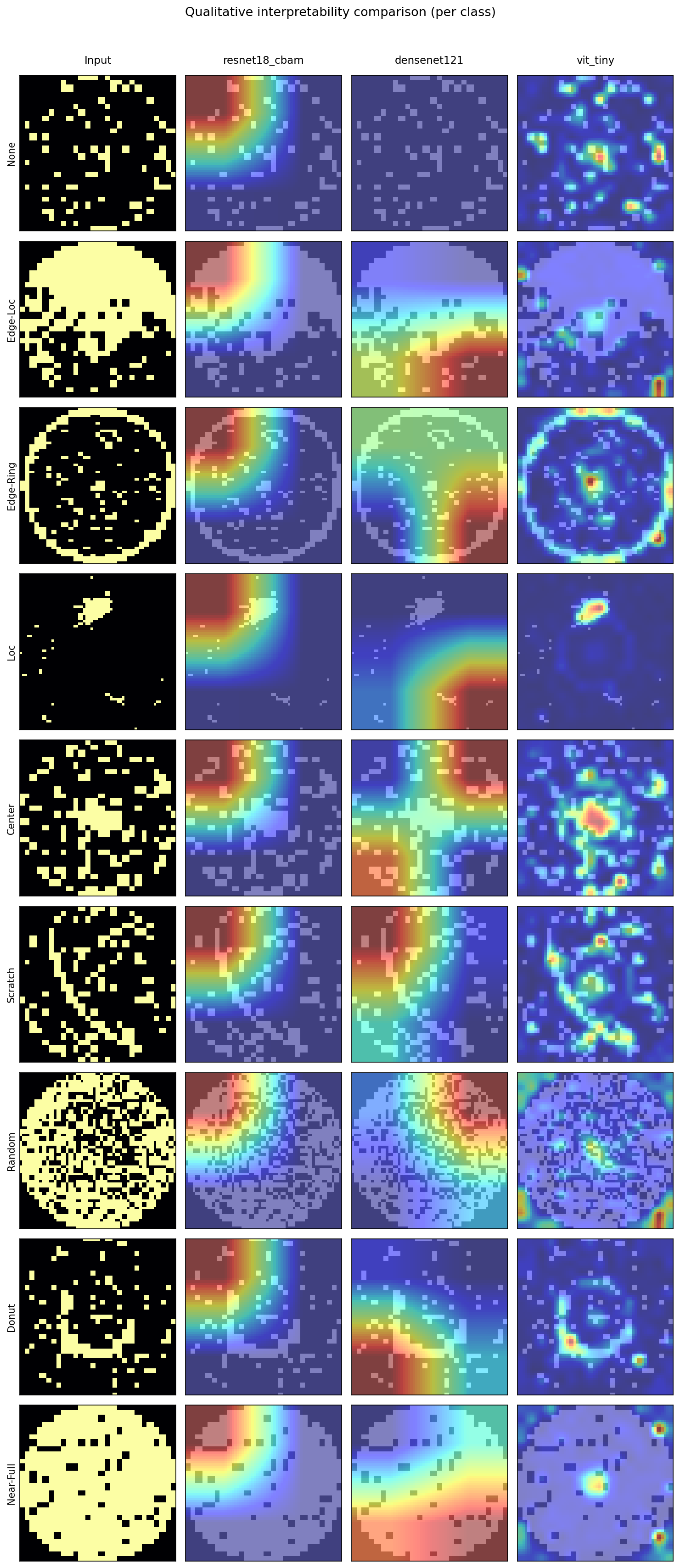}
\caption{\textbf{Figure 5.} Qualitative heatmap comparison (highest-confidence correct sample per class)}
\end{figure}

Note that Figures 4a--4c and Figure 5 use different samples (the close-ups come from the first correctly-classified test sample per run, while the panel selects the highest-confidence sample across seeds). Figure 4a illustrates a degenerate Grad-CAM case --- not atypical for ResNet18+CBAM on certain inputs --- whereas the panel shows the more common quadrant-locked pattern. (Degenerate maps arise when the gradient-weighted activation after ReLU has near-zero positive mass for the target class; min--max normalisation cannot then recover a meaningful spatial ranking.)

\textbf{Grad-CAM sanity checks.} To rule out implementation artefacts, we verified that: (i) gradients were non-zero at the target layers (\texttt{cbam4} for ResNet18+CBAM, \texttt{features.denseblock4} for DenseNet121); (ii) backpropagation was performed with respect to the target logit (pre-softmax), not the softmax probability; (iii) heatmaps were computed with respect to the predicted class; and (iv) in pilot checks, removing the Grad-CAM ReLU (allowing negative contributions) did not materially change the family-level Deletion AUC ranking. The degenerate behaviour is therefore a property of the gradient signal at these layers, not an implementation error.

The three methods produce qualitatively distinct explanations despite operating on models trained with identical data and optimisation protocols. A note on interpretation: Figure 5 displays each heatmap \emph{overlaid} on the input wafer (semi-transparent), which can visually blur the intrinsic behaviour of the heatmap with the spatial structure of the wafer itself. The raw heatmaps --- with the wafer background removed and each map normalised to its own dynamic range --- are provided in Supplementary Figure S2 and reveal a markedly cleaner picture of each method's behaviour.

ResNet18+CBAM Grad-CAM yields broad, smoothly varying heatmaps that are \textbf{essentially input-independent}: in the raw view (Supplementary Figure S2), a near-identical red blob occupies the upper-left corner of every wafer across all nine classes, with only minor variation in the intensity contour. The apparent class-dependent structure in the overlay is largely caused by the wafer background showing through a nearly fixed Grad-CAM blob, so the map behaves more like a coarse quadrant-level activation pattern than a shape-aware defect localisation.

DenseNet121 Grad-CAM produces heatmaps of comparable spatial extent but places them in a different --- and also largely input-independent --- region: most classes attract a lower-right or lower-left blob that again does not trace the defect geometry. The two Grad-CAMs rarely agree on quadrant for the same input, even though both underlying models achieve essentially identical classification accuracy (balanced accuracy $\approx$ 0.901). This architecture-dependent disagreement is informative: it indicates that Grad-CAM reports model-specific gradient paths rather than a unique ground-truth saliency, and that the apparent "attention" of a CNN to a region is largely a property of its learned filter statistics, not of the input.

This qualitative impression is corroborated quantitatively by the Spearman rank correlation between each heatmap and the wafer's defect-pixel distribution (Supplementary Table S4): ResNet18+CBAM achieves a mean correlation of 0.037 $\pm$ 0.105, DenseNet121 0.015 $\pm$ 0.126 --- both statistically indistinguishable from zero --- while ViT-Tiny reaches 0.318 $\pm$ 0.209, roughly 8--21$\times$ higher. In other words, the two Grad-CAMs carry essentially no spatial information about where defects lie, whereas Attention Rollout carries a measurable (if noisy) spatial signal.

ViT-Tiny Attention Rollout departs from both CNN families. Rather than painting broad regions, it produces sparse, high-contrast hot-spots whose locations vary meaningfully with the input (Supplementary Figure S2). The spatial correspondence with defects is strongest for classes with a single compact, geometrically distinctive signature: in \textbf{Center} the main hot-spot sits directly on the central defect cluster, in \textbf{Loc} on the localised cluster, and in \textbf{Edge-Ring} the hot-spots trace the ring shape itself (distributed around the circumference rather than concentrated on one arc). For other classes the alignment is weaker: in \textbf{Donut} the hot-spots scatter without cleanly outlining the ring; in \textbf{Edge-Loc} they partially track the edge but with substantial off-target activation; in \textbf{Near-Full} a central hot-spot remains despite the defect pattern
being near-global. For \textbf{None}, the heatmap lacks a coherent defect-shaped structure, as expected given the absence of labelled defect signal. For \textbf{Random}, hot-spots scatter across the wafer without tracking the diffuse defect pixels, though the activation pattern is visually distinguishable from None --- a human reviewer could use this contrast even though the method does not localise the defect. \textbf{Scratch} partially tracks the arc-shaped defect trajectory, performing somewhat better than Edge-Loc in spatial alignment despite the fine-grained nature of the pattern. Across most classes a recurring set of boundary hot-spots also appears near the circular wafer edge; these likely reflect the model attending to the background/wafer transition rather than class-specific evidence. Critically, even when ViT-Tiny's spatial localisation is imperfect, its \textbf{per-sample}
hot-spot pattern changes with the input, which is qualitatively distinct from the quadrant-locked, essentially class-invariant behaviour of the two Grad-CAMs. The fine-grained localisation is visually compelling, yet, as shown in \S{}4.1, ViT-Tiny's classification accuracy is lower than either CNN family by roughly 10 percentage points on macro-F1 --- an observation that raises the central question: is the ViT heatmap faithful to ViT-Tiny's weaker decisions, or is it merely prettier than the CNN heatmaps? The quantitative faithfulness analysis in \S{}4.3 adjudicates this question.

\subsection{Quantitative faithfulness (aggregate)}

The three faithfulness metrics defined in \S{}3.4.3 are reported in Table 4, averaged across 594 samples per family (200 balanced-stratified samples per seed $\times$ 3 seeds). ViT-Tiny with Attention Rollout achieves the lowest (most faithful) Deletion AUC --- less than half that of either CNN family --- and correspondingly the highest Insertion AUC. Stability is comparable across families in mean (0.834--0.891), but ViT-Tiny exhibits the tightest distribution (standard deviation 0.107 versus 0.258--0.275 for the CNN families).

\begin{table}[htbp]
\centering
\small
\caption{\textbf{Table 4.} Quantitative faithfulness metrics (mean $\pm$ std, n=594)}
\resizebox{\linewidth}{!}{%
\begin{tabular}{lrrrrr}
\toprule
Family & N runs & N samples & Deletion AUC $\downarrow$ & Insertion AUC $\uparrow$ & Stability $\uparrow$ \\
\midrule
ResNet18+CBAM & 3 & 594 & 0.495 $\pm$ 0.270 & 0.534 $\pm$ 0.247 & 0.834 $\pm$ 0.275 \\
DenseNet121 & 3 & 594 & 0.525 $\pm$ 0.286 & 0.511 $\pm$ 0.268 & 0.891 $\pm$ 0.258 \\
ViT-Tiny & 3 & 594 & 0.211 $\pm$ 0.233 & 0.693 $\pm$ 0.274 & 0.845 $\pm$ 0.107 \\
Swin-Tiny & 3 & 594 & 0.432 $\pm$ 0.233 & 0.480 $\pm$ 0.222 & 0.893 $\pm$ 0.097 \\
\bottomrule
\end{tabular}
}
\end{table}

The mean predicted-class probability as pixels are progressively removed (deletion) or added (insertion) is shown in Figure 6. ViT-Tiny's deletion curve drops steeply within the first 20 \% of pixels removed, reaching near-chance levels by 40 \%, indicating that Attention Rollout concentrates importance on a compact set of prediction-relevant pixels. Both CNN families decline more gradually, indicating that their Grad-CAM heatmaps spread importance over a larger area that includes many non-essential pixels. On the insertion side, ViT-Tiny recovers more than 60 \% confidence after inserting just 40 \% of pixels, while the CNN families require nearly all pixels to reach comparable confidence.

\begin{figure}[htbp]
\centering
\includegraphics[width=0.98\linewidth]{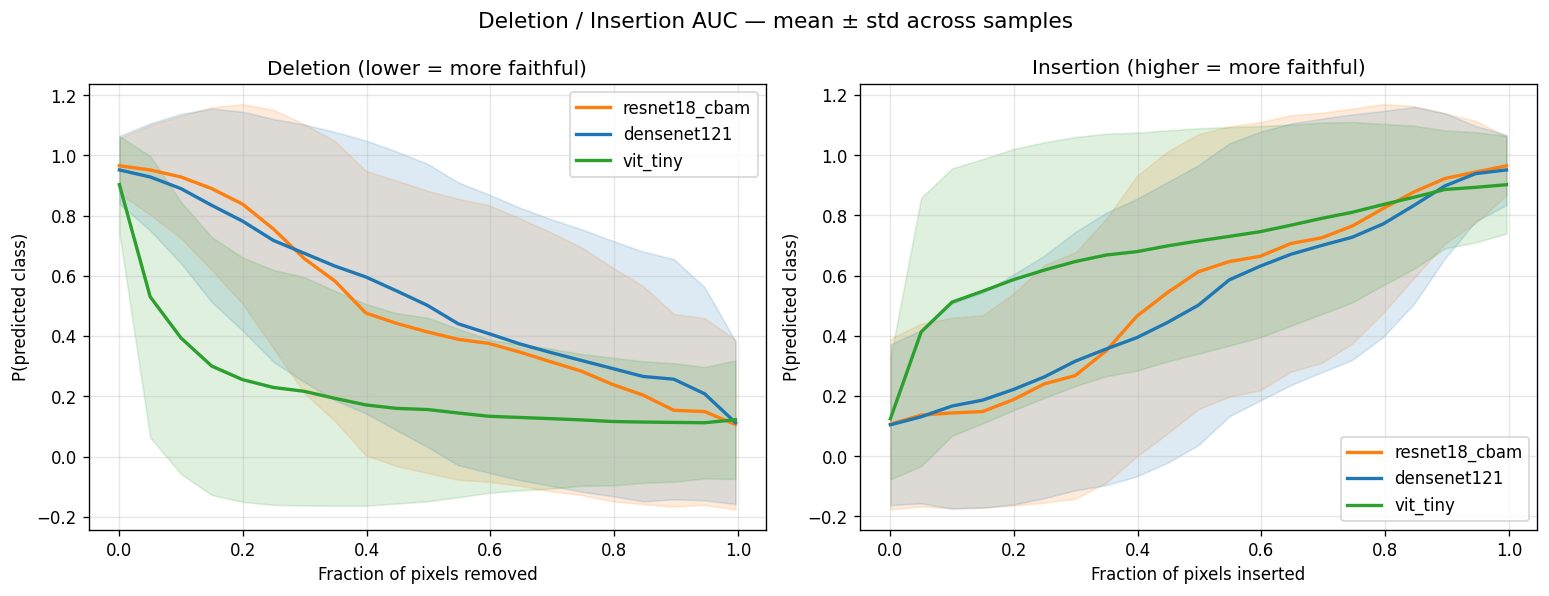}
\caption{\textbf{Figure 6.} Deletion and Insertion curves (mean $\pm$ std across 594 samples)}
\end{figure}

Per-sample stability --- the cosine similarity of the heatmap under the $K{=}5$ semantics-preserving perturbations defined in \S{}3.4.3 --- is reported as a boxplot in Figure 7. DenseNet121 attains the highest median stability (approximately 0.99) but also the widest spread, with a cluster of outliers near zero. These outliers correspond predominantly to "None"-class samples for which Grad-CAM produces degenerate all-zero heatmaps; in that regime any small perturbation yields an essentially unrelated heatmap, collapsing the cosine similarity to zero. ResNet18+CBAM exhibits a qualitatively similar profile: a high median accompanied by a small number of low-stability outliers driven by the same degenerate-heatmap mechanism. ViT-Tiny presents a different pattern. Its median stability is slightly lower (approximately 0.93), but its interquartile range is markedly tighter (standard deviation
0.107 versus 0.258--0.275 for the CNN families), indicating that its explanations are uniformly robust across all classes --- including "None" --- because Attention Rollout always produces a non-trivial distribution over patches.

An alternative interpretation merits acknowledgment: on non-degenerate samples, DenseNet's high stability could indicate that its Grad-CAM heatmaps, while less faithful in the deletion sense, are at least \emph{consistently} unfaithful --- always highlighting the same prediction- irrelevant region. An equivalent interpretation is that ViT-Tiny's lower stability reflects its tighter coupling to defect location: augmentations that shift defect pixels necessarily shift the heatmap, whereas CNN heatmaps are stable precisely because they are insensitive to the spatial details that actually drive the prediction. Stability alone therefore does not imply faithfulness; it must be read in conjunction with the Deletion and Insertion metrics above.

\begin{figure}[htbp]
\centering
\includegraphics[width=0.60\linewidth]{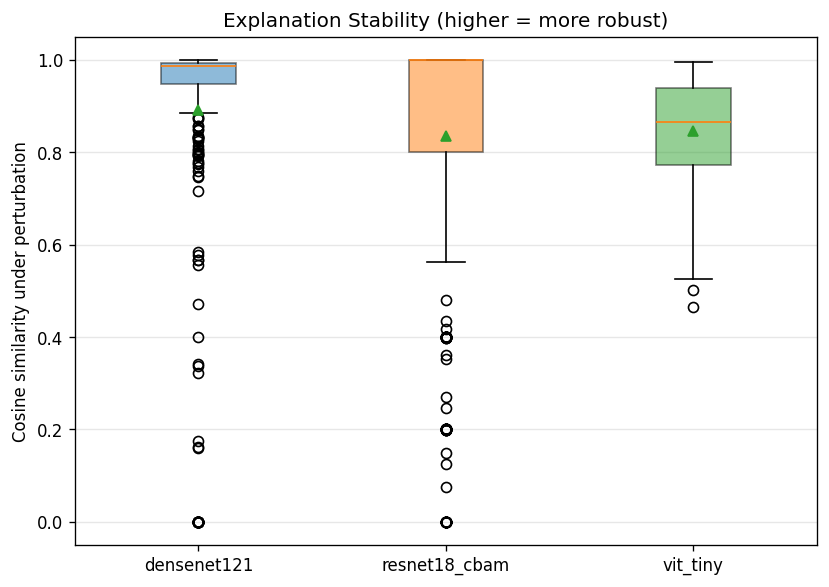}
\caption{\textbf{Figure 7.} Explanation stability distribution (per-sample cosine similarity across K=5 perturbations)}
\end{figure}

This pattern is best read as evidence for a composite architecture--explainer distance. Readout directness and spatial granularity co-vary across the four families: ViT-Tiny is close on both axes, Swin-Tiny is intermediate, and the two CNN families are farther on both. The ViT Grad-CAM ablation (\S{}S9) and final-layer CLS attention ablation (\S{}S10) provide a partial decomposition, suggesting that both direct access to native attention and multi-layer rollout depth contribute materially to the observed gap.

\subsection{Per-class breakdown, effect size, and bootstrap confidence intervals}

The aggregate family-level comparison reported above summarises three numbers per metric. This subsection disaggregates the same measurements along two directions --- across the nine defect classes and across resampled bootstrap replicates --- to assess whether the native-readout advantage is a property of the dataset as a whole or an artefact of a favourable subset.

Table 5 reports the mean Deletion AUC for each (family, class) cell. ViT-Tiny achieves the lowest (most faithful) Deletion AUC in \textbf{eight of the nine classes}, the sole exception being the \emph{Near-full} class where the defect covers almost the entire wafer and the faithfulness metric becomes degenerate for all families. In this class, DenseNet121 achieves the lowest Deletion AUC (0.151), which is mechanistically expected: when the defect is spatially uniform across the wafer, the coarse receptive field of Grad-CAM is no longer a disadvantage because there is no fine-grained spatial structure to resolve. This class-level consistency is difficult to attribute to noise: were the native-readout advantage present only at the aggregate level, one would expect the ordering to invert in several individual classes.

Table 6 reports the pooled effect size and 95 \% bootstrap confidence intervals for the family-level Deletion AUC. Cohen's \emph{d} between ViT-Tiny and each CNN family exceeds 1.1 in absolute magnitude, which is conventionally interpreted as a \emph{very large} effect. The 95 \% bootstrap confidence intervals, computed by resampling the 594 per-sample measurements per family 2,000 times, do not overlap: ViT-Tiny lies in [0.192, 0.231], Swin-Tiny in [0.413, 0.450], ResNet18+CBAM in [0.474, 0.515], and DenseNet121 in [0.502, 0.548]. Thus, ViT-Tiny is separable from all three Grad-CAM families at the evaluated sample-pool level. Swin-Tiny is also separable from the two CNN families, but remains much closer to them than to ViT-Tiny, consistent with its intermediate readout-structure position. Because the 594 measurements are nested within only three trained seeds, these bootstrap intervals should be interpreted as uncertainty over the evaluated sample pool rather than as a full estimate of experiment-level uncertainty across independently trained populations; the seed-level means reported in Supplementary Table S5 provide the complementary seed-level view.

A final robustness check addresses the concern that the aggregate result might be driven by the over-represented "None" class (66 of 594 samples per family under balanced-stratified sampling). Excluding all None-class samples and recomputing the mean Deletion AUC yields 0.163 for ViT-Tiny, 0.466 for ResNet18+CBAM, and 0.482 for DenseNet121. The family ordering is unchanged and the ViT-Tiny margin over the CNN families is comparable to the full-sample margin (approximately 0.30 on defect classes versus approximately 0.28--0.31 on all classes). The aggregate effect is therefore not an artefact of the majority class.

\begin{table}[htbp]
\centering
\small
\caption{\textbf{Table 5.} Per-class Deletion AUC (lower = more faithful). ViT-Tiny is most faithful in 8/9 classes}
\resizebox{\linewidth}{!}{%
\begin{tabular}{lrrrrrrrrr}
\toprule
Family & None & Edge-Loc & Edge-Ring & Loc & Center & Scratch & Random & Donut & Near-Full \\
\midrule
DenseNet121 & 0.873 & 0.712 & 0.641 & 0.745 & 0.476 & 0.428 & 0.318 & 0.384 & 0.151 \\
ResNet18+CBAM & 0.719 & 0.644 & 0.762 & 0.548 & 0.367 & 0.596 & 0.309 & 0.312 & 0.195 \\
Swin-Tiny & 0.609 & 0.428 & 0.553 & 0.619 & 0.411 & 0.440 & 0.282 & 0.357 & 0.190 \\
ViT-Tiny & 0.593 & 0.157 & 0.094 & 0.246 & 0.073 & 0.348 & 0.105 & 0.091 & 0.189 \\
\bottomrule
\end{tabular}
}
\end{table}

\begin{table}[htbp]
\centering
\small
\caption{\textbf{Table 6.} Bootstrap 95 \% confidence intervals (2,000 resamples) and Cohen's \emph{d} for pooled Deletion AUC over the evaluated sample set. Because samples are nested within only three independently trained seeds, these intervals describe uncertainty over the sample pool and should not be interpreted as a substitute for a hierarchical or seed-level uncertainty estimate. $|d| > 1.1$ indicates a large effect}
\begin{tabular}{lrrrr}
\toprule
Family & Mean Del AUC & Mean (excl. None) & 95\% CI & n \\
\midrule
ViT-Tiny & 0.211 & 0.163 & [0.192, 0.231] & 594 \\
Swin-Tiny & 0.432 & 0.410 & [0.413, 0.450] & 594 \\
ResNet18+CBAM & 0.495 & 0.466 & [0.474, 0.515] & 594 \\
DenseNet121 & 0.525 & 0.482 & [0.503, 0.548] & 594 \\
\bottomrule
\end{tabular}
\end{table}

\begin{Verbatim}
Cohen d (ViT vs Swin-Tiny): approximately -0.95  [large effect]
Cohen d (ViT vs ResNet+CBAM): -1.13  [very large effect]
Cohen d (ViT vs DenseNet121): -1.21  [very large effect]
\end{Verbatim}

\subsection{Model-agnostic control: RISE on WM-811K}

The fourth prediction of the native-readout hypothesis (\S{}1.2) states that a model-agnostic explainer should greatly compress the faithfulness ranking across families. Table 7 reports RISE alongside the native methods on WM-811K.

\begin{center}
\small \textbf{Table 7.} Native vs. RISE faithfulness on WM-811K (primary seed per family: seed 7 for Swin-Tiny, seed 42 for others; n=198 samples). Deletion AUC: lower is more faithful; Insertion AUC: higher is more faithful
\end{center}

\begin{tabular}{lcccc}
\toprule
Family & Native Del $\downarrow$ & RISE Del $\downarrow$ & Native Ins $\uparrow$ & RISE Ins $\uparrow$ \\
\midrule
ResNet18+CBAM & 0.486 & \textbf{0.091} & 0.531 & \textbf{0.858} \\
DenseNet121 & 0.544 & \textbf{0.130} & 0.501 & \textbf{0.823} \\
Swin-Tiny & 0.432 & \textbf{0.096} & 0.480 & \textbf{0.823} \\
ViT-Tiny & \textbf{0.221} & \textbf{0.093} & \textbf{0.689} & \textbf{0.818} \\
\bottomrule
\end{tabular}

The result is notable: RISE greatly compresses the Deletion AUC spread from 0.221--0.544 (native, primary seed) to 0.091--0.130 (RISE). Under native methods, ViT-Tiny dominates with Del 0.221 and Swin-Tiny is intermediate at 0.432; under RISE, all four families achieve Del < 0.13 with a greatly compressed spread. This is consistent with the fourth prediction: the faithfulness gap observed with native methods resides in the \emph{explainer pathway}, not in the architecture's learned representations. When the explainer is held constant (RISE for all), the native-method gap is largely removed, and the architectures appear similarly probeable by RISE under this WM-811K protocol. Swin-Tiny's RISE performance (Del 0.096) is virtually identical to ViT-Tiny (0.093) and ResNet18+CBAM (0.091), consistent with the interpretation that its intermediate native-method Deletion AUC (0.432) is attributable to the Grad-CAM explainer pathway rather than to any inherent limitation of its learned representations. Note that Swin's advantage over the CNNs is metric-dependent: while Deletion AUC improves (0.432 vs 0.495--0.525), Insertion AUC does not (0.480 vs 0.511--0.534), suggesting that Swin's spatial hierarchy benefits pixel-removal sensitivity more than pixel-insertion recovery.

Notably, RISE achieves \emph{better} perturbation-based faithfulness than any native method under this protocol (Del 0.09 vs native-best 0.22). This does not contradict the hypothesis --- it indicates that RISE, by directly probing the model's input--output function, can identify decision-relevant pixels that gradient-based or attention-based readouts miss. The hypothesis concerns \emph{relative} ordering across families, not absolute performance.

From a practical standpoint, if the sole objective is maximising perturbation-based faithfulness and computational cost is acceptable, RISE (4,000 forward passes per sample) may be preferable as an offline audit reference. Native readout remains relevant when (i) computational budget is limited (a single forward pass for Rollout vs. 4,000 for RISE), (ii) explanations must be generated interactively for operator-facing review, or (iii) the goal is to understand \emph{why} faithfulness varies across architecture--explainer pairings rather than merely to maximise it.

\subsection{Perturbation-baseline sensitivity}

The preceding WM-811K results use zero-fill as the perturbation baseline --- a domain-motivated choice because wafer maps are sparse discrete die-state arrays where zero corresponds to the background/non-wafer value. To test whether the family ordering is robust to this choice, we repeat the native-method Deletion evaluation using Gaussian blur-fill ($\sigma$ = 3.0) as an alternative baseline.

\begin{center}
\small \textbf{Table 8.} Deletion AUC under zero-fill vs blur-fill (WM-811K, primary seed per family, n = 198)
\end{center}

\begin{tabular}{lcc}
\toprule
Family & Zero-fill Del $\downarrow$ & Blur-fill Del $\downarrow$ \\
\midrule
ViT-Tiny (Rollout) & \textbf{0.221} & 0.693 \\
Swin-Tiny (Grad-CAM) & 0.440 & 0.700 \\
ResNet18+CBAM (Grad-CAM) & 0.486 & \textbf{0.559} \\
DenseNet121 (Grad-CAM) & 0.544 & 0.681 \\
\bottomrule
\end{tabular}

Under blur-fill, the ViT advantage disappears and the family ordering changes substantially: ResNet18+CBAM becomes best, while ViT-Tiny and Swin-Tiny become the two weakest families, with Swin-Tiny numerically worst by a small margin (0.700 vs. 0.693). All families converge to a narrower range (0.56--0.70). Swin-Tiny's zero-fill intermediate position (0.440) therefore shifts to the bottom of the blur-fill ranking, indicating that sensitivity to the fill operator is not limited to the global-attention ViT. This suggests that the zero-fill ViT advantage arises partly from an interaction between sparse patch-level heatmaps and the zero-fill operator, which removes local evidence completely rather than smoothing it.

This result does not invalidate the native-readout hypothesis but qualifies its scope: the hypothesis predicts relative faithfulness under a specified perturbation protocol. For discrete wafer maps, zero-fill is the primary domain-motivated evidence-removal operation because setting selected pixels to the background value removes local die-state evidence. Blur-fill tests a different perturbation semantics. Instead of removing evidence, it smooths selected regions using neighboring pixels; for compact defect clusters, this may partially reconstruct or preserve the defect signal. Consequently, a sparse and well-localized heatmap such as ViT Rollout can be penalized under blur-fill because replacing its top-ranked pixels with blurred neighbors may leave enough defect evidence for the classifier to maintain confidence. Therefore, blur-fill sensitivity shows that perturbation-based
faithfulness depends on the fill operator and should be reported as part of the audit protocol.

\subsection{Boundary-condition study: MVTec AD (exploratory)}

MVTec AD serves as an exploratory boundary-condition study rather than a fully matched replication or generalization proof: models are pretrained on ImageNet, the task is binary (defective vs good), and only one seed is evaluated. The results are therefore used to identify where the WM-811K findings require qualification, not to claim transfer of the same ranking to all industrial anomaly-detection settings.

\subsubsection{Classification performance}

Table 9 reports classification performance on MVTec AD (binary: good vs. defective).

\begin{center}
\small \textbf{Table 9.} MVTec AD classification performance (seed 42)
\end{center}

\begin{tabular}{lccc}
\toprule
Family & Accuracy & Balanced Acc & F1 (macro) \\
\midrule
DenseNet121 & 0.970 & 0.944 & 0.957 \\
Swin-Tiny (pretrained) & 0.963 & 0.937 & 0.947 \\
ResNet18+CBAM & 0.955 & 0.919 & 0.935 \\
ViT-Tiny (pretrained) & 0.917 & 0.872 & 0.881 \\
\bottomrule
\end{tabular}

All families achieve strong classification performance. The three spatial-hierarchy models (DenseNet, Swin, ResNet) cluster at 92--94 \% balanced accuracy, while the global-attention model (ViT-Tiny) is lower (87 \%). Notably, Swin-Tiny --- a transformer --- performs comparably to the CNNs, indicating that its hierarchical structure provides sufficient inductive bias for this task. This performance gap is acknowledged as a confound in the Limitations section (\S{}5.7).

\subsubsection{Interpretability: native methods vs. RISE}

Table 10 reports faithfulness metrics on MVTec AD (200 defective test samples, seed 42). IoU is computed against pixel-level ground-truth defect masks.

\begin{center}
\small \textbf{Table 10.} MVTec AD interpretability (seed 42, n=200). IoU vs ground-truth masks
\end{center}

\noindent\resizebox{\linewidth}{!}{%
\begin{tabular}{lcccccc}
\toprule
Family & Native Del $\downarrow$ & RISE Del $\downarrow$ & Native Ins $\uparrow$ & RISE Ins $\uparrow$ & Native IoU $\uparrow$ & RISE IoU $\uparrow$ \\
\midrule
ResNet18+CBAM & \textbf{0.413} & 0.886 & 0.846 & 0.905 & 0.133 & 0.165 \\
Swin-Tiny & 0.509 & 0.709 & 0.909 & 0.927 & 0.102 & 0.162 \\
DenseNet121 & 0.671 & 0.938 & \textbf{0.944} & 0.937 & 0.141 & 0.206 \\
ViT-Tiny & 0.770 & \textbf{0.656} & 0.855 & 0.792 & \textbf{0.244} & 0.116 \\
\bottomrule
\end{tabular}
}

The MVTec results reveal a clear pattern when Swin-Tiny is included:

\textbf{(1) Spatial-hierarchy models have native methods that beat RISE.} For ResNet (0.413 vs 0.886), Swin (0.509 vs 0.709), and DenseNet (0.671 vs 0.938), native Grad-CAM achieves substantially lower (better) Deletion AUC than RISE. Only ViT-Tiny reverses this pattern (native 0.770 vs RISE 0.656). This supports Prediction 5: Swin behaves like the CNNs, not like ViT, because its spatial hierarchy provides a compatible readout path for Grad-CAM.

\textbf{(2) RISE does not compress.} RISE Deletion AUC ranges from 0.656 to 0.938 --- a wide spread with no convergence, unlike WM-811K where RISE compressed all families to Del < 0.13.

\textbf{(3) ViT-Tiny has the highest ground-truth IoU (0.244).} Despite the worst Deletion AUC, ViT's attention maps best localise actual defect pixels. This dissociation between faithfulness-to-model and faithfulness-to-ground-truth is discussed in \S{}5.4.

\subsection{RISE mask-resolution ablation (MVTec AD)}

To test whether RISE's poor performance on MVTec AD is an artefact of insufficient mask resolution, the evaluation is repeated at mask resolutions 16 and 32 (cell sizes 16 px and 8 px respectively).

\begin{center}
\small \textbf{Table 11.} RISE Deletion AUC on MVTec AD as a function of mask resolution (mask\_res=8: n=200; mask\_res=16,32: n=50 samples)
\end{center}

\begin{tabular}{lccc}
\toprule
Family & mask\_res=8 & mask\_res=16 & mask\_res=32 \\
\midrule
ResNet18+CBAM & 0.886 & 0.893 & 0.839 \\
Swin-Tiny & 0.709 & 0.855 & 0.795 \\
DenseNet121 & 0.938 & 0.960 & 0.939 \\
ViT-Tiny & 0.656 & 0.989 & 0.914 \\
\bottomrule
\end{tabular}

Increasing mask resolution does not systematically improve RISE's faithfulness --- in fact, performance degrades for ViT-Tiny (0.656 $\rightarrow$ 0.989 at mask\_res=16) and Swin-Tiny (0.709 $\rightarrow$ 0.855). This rules out the tested mask-resolution explanation: RISE's poor performance on MVTec AD is not explained by the coarse 8$\times$8 mask grid alone, though other RISE hyperparameters (number of masks, fill value, smoothing) remain untested. Swin-Tiny's pattern mirrors the CNNs (RISE consistently worse than native Grad-CAM at all resolutions), further supporting its alignment with spatial-hierarchy models. The default mask\_res=8 is already near-optimal for WM-811K (64$\times$64 images, cell = 8 px $\approx$ defect scale); a corresponding WM-811K mask-resolution ablation is left for future sensitivity analysis.

\section{Discussion}

\subsection{Testing the native-readout hypothesis}

The following interpretation concerns model--explainer pairs, not architectures in isolation. The five predictions set out in \S{}1.2 can now be evaluated against the measurements reported in \S{}4.3--4.6, with the important qualification that the primary claim is protocol-conditioned. The first prediction stated that explanations reading natively from a model's decision operators should produce lower Deletion AUC and higher Insertion AUC than explanations that approximate those operators post-hoc under the specified zero-fill protocol. The Deletion AUC of ViT-Tiny + Rollout (0.21) is less than half of any Grad-CAM combination (0.43--0.53), and the Insertion AUC ranking is correspondingly inverted (0.693 vs 0.480--0.534), consistent with the prediction. The second prediction stated that faithfulness ranking should be insensitive to classification ranking; \S{}5.2 establishes this dissociation directly. The third prediction stated that faithfulness ranking should be stable across random seeds. The per-seed
mean Deletion AUC values are reported in Supplementary Table S5; the between-seed standard deviation of each family's mean Deletion AUC is under 0.02 for all four families (Supplementary Table S5), and the family ordering --- ViT-Tiny most faithful, then Swin-Tiny, then ResNet18+CBAM, then DenseNet121 --- is unambiguous at every seed.

The fourth prediction --- that a model-agnostic explainer should compress the faithfulness ranking --- receives support from the RISE experiment (\S{}4.5): applying the same model-agnostic explainer to all families compresses the Deletion AUC range from 0.211--0.525 under native methods to 0.091--0.130. This indicates that the native-method gap is primarily a property of the explainer pathway under the WM-811K zero-fill protocol, rather than a difference in whether the architectures are probeable by perturbation methods.

The fifth prediction --- that a hierarchical transformer should align with Grad-CAM rather than Attention Rollout --- receives converging support from both datasets. On WM-811K, Swin-Tiny's Grad-CAM achieves Deletion AUC 0.432, placing it between ViT (0.211) and the CNNs (0.495--0.525) rather than matching ViT's Rollout performance. On MVTec AD, Swin's native Grad-CAM beats RISE on Deletion (0.509 vs 0.709), the same pattern as both CNNs and the opposite of ViT (where RISE beats native). This supports the interpretation that the operative factor for Grad-CAM compatibility is \textbf{spatial hierarchy} --- the presence of progressively- downsampled feature maps that preserve spatial structure --- rather than the convolution/attention distinction. Swin is a transformer in every computational sense (self-attention is its sole primitive), yet its hierarchical spatial structure makes Grad-CAM
a compatible readout, just as it is for CNNs.

Two structural proxies for "distance from the native decision mechanism" operationalise the hypothesis. The first is \textbf{spatial granularity}: ViT-Tiny reads importance at 4$\times$4 patch resolution (16 pixels per token, approximately 0.4 \% of the 64$\times$64 image), whereas Grad-CAM reads importance through the effective receptive field of \texttt{cbam4}, \texttt{denseblock4}, or Swin's \texttt{final\_stage}, which operates at a substantially coarser spatial scale than the 4$\times$4 ViT patch grid. A method that can only assign importance at coarse spatial scale is structurally prevented from surgically removing the decision-relevant pixels under a Deletion perturbation. The second proxy is \textbf{readout directness}: Rollout reads attention matrices that form part of the model's native information-routing pathway to the classifier, while Grad-CAM reconstructs importance from
gradients of activations never intended to expose an interpretable routing signal. Both proxies predict the observed ordering, and neither is a property of the specific training run --- they are architectural properties of the model--explainer pair.

Swin-Tiny's intermediate position (Del 0.432, between ViT's 0.211 and the CNNs' 0.495--0.525) is consistent with this framework. Swin preserves an explicit hierarchical spatial representation throughout the network, making Grad-CAM structurally applicable to its final stage. However, because the final-stage feature map is coarse (2$\times$2 for 64$\times$64 input), Swin remains less spatially precise than ViT Rollout on the 4$\times$4 patch grid. Its readout is indirect (Grad-CAM, not attention matrices), placing it closer to the CNNs on the directness axis. The key property is not finer final-layer resolution but the preservation of spatial hierarchy through attention-based downsampling, which makes Grad-CAM structurally compatible. The net result is an intermediate Deletion AUC --- exactly as the two-proxy framework predicts.

A methodological caveat applies to the attribution of the observed effect between the two proxies. Across the four families examined here, the two proxies are largely rank-correlated: ViT-Tiny has minimal structural distance on both (direct readout, fine granularity), Swin-Tiny is intermediate (indirect readout, moderate granularity), while both CNN families have large structural distance on both (indirect readout, coarse granularity). Two ViT-specific ablations partially disentangle readout directness from rollout depth, while leaving spatial granularity only indirectly controlled: \textbf{(i) Grad-CAM applied to ViT-Tiny} (Supplementary Table S7), which holds spatial granularity approximately constant while increasing readout distance --- the result (Del 0.492, indistinguishable from CNN Grad-CAM) indicates that readout directness carries substantial signal; and \textbf{(ii) final-layer CLS attention on ViT-Tiny} (Supplementary Table S10), which holds readout origin constant (native
attention) while removing multi-layer rollout depth --- the result (Del 0.367 $\pm$ 0.031 across three seeds) falls between full Rollout (0.211) and Grad-CAM-on-ViT (0.492). Together, these two ablations decompose the total Rollout-vs-CNN gap ($\approx$0.30) into approximately 42 \% attributable to readout directness and 52 \% to rollout depth, with both manipulated factors carrying independent signal (the residual $\approx$6 \% reflects non-additivity of the two manipulations; the partition is path-dependent). The composite structural distance therefore predicts faithfulness, and neither proxy alone accounts for the full effect.

An additional line of evidence supports interpreting the effect as architectural rather than noise-driven. Per-class analysis (\S{}4.4) shows ViT-Tiny achieving the lowest Deletion AUC in 8 of 9 defect classes, not only in the pooled mean. The pooled effect size is Cohen's \emph{d} $\approx$ 1.1--1.2 against either CNN family, with non- overlapping 95 \% bootstrap confidence intervals. The WM-811K model--explainer-pair difference is therefore separable at the evaluated sample-pool level, consistent with a structural rather than a statistical-accident explanation.

Taken together, the results support an audit-protocol interpretation rather than a universal explainer ranking. The WM-811K zero-fill results show that architecture--explainer compatibility can strongly affect perturbation faithfulness. The RISE results show that native readout is not an optimality guarantee: model-agnostic perturbation probing can achieve stronger perturbation faithfulness and serves as a strong offline audit reference. The blur-fill sensitivity analysis further shows that measured faithfulness depends on the perturbation operator. Therefore, explanation faithfulness in industrial visual inspection should be treated as an empirical property of a model--explainer--perturbation protocol. Heatmaps should not be deployed solely because they are visually plausible or architecture-native; they should be audited under a documented perturbation protocol and compared against
model-agnostic baselines.

\subsection{Classification versus interpretability}

Table 12 juxtaposes the primary classification metric (macro-F1) against the two principal faithfulness metrics (Deletion AUC and Insertion AUC), exposing a clear dissociation between the two dimensions of model quality. DenseNet121 and ResNet18+CBAM attain nearly identical macro-F1 scores of 0.853 and 0.849, respectively, while Swin-Tiny achieves 0.792 and ViT-Tiny trails at 0.747. When the same models are ranked by Deletion AUC --- lower being more faithful --- the order inverts: ViT-Tiny achieves 0.211, whereas ResNet18+CBAM reaches 0.495 and DenseNet121 0.525. The Insertion AUC comparison yields the same ordering, with ViT-Tiny at 0.693 ahead of ResNet18+CBAM (0.534) and DenseNet121 (0.511). The convolutional families therefore dominate classification, whereas the global-attention ViT paired with Rollout dominates perturbation-based faithfulness. This dissociation supports the second
prediction of the native-readout hypothesis (\S{}1.2): the faithfulness ranking of families is insensitive to their classification ranking, because the two quantities measure structurally independent properties of the model--explainer pair.

\begin{table}[htbp]
\centering
\small
\caption{\textbf{Table 12.} Classification versus faithfulness on WM-811K (F1 Macro: 3-seed mean; faithfulness metrics: pooled across 594 samples)}
\begin{tabular}{lrrr}
\toprule
Family & F1 Macro & Deletion AUC $\downarrow$ & Insertion AUC $\uparrow$ \\
\midrule
DenseNet121 & 0.853 (best) & 0.525 (worst) & 0.511 (worst) \\
ResNet18+CBAM & 0.849 & 0.495 & 0.534 \\
Swin-Tiny & 0.792 & 0.432 & 0.480 \\
ViT-Tiny & 0.747 (worst) & 0.211 (best) & 0.693 (best) \\
\bottomrule
\end{tabular}
\end{table}

\subsection{Does visual coherence imply faithfulness?}

The evidence assembled in \S{}4.2 and \S{}4.3 indicates that it does not, and the native-readout framework supplies a mechanistic account of why. Qualitatively, CNN Grad-CAM heatmaps appear smooth and plausible, highlighting broad regions that are expected to coincide with the defect area; quantitatively, however, their Deletion AUC of approximately 0.5 reveals that a substantial fraction of those highlighted pixels can be zeroed out without materially affecting the prediction. A random-pixel baseline (Supplementary Table S6) shows that CNN Grad-CAM performs no better than random ordering (ResNet Del 0.486 vs random 0.498; DenseNet Del 0.544 vs random 0.508), whereas ViT-Tiny Rollout (Del 0.221) is substantially better than the random baseline (0.480). ViT-Tiny's sparse, high-contrast hot-spots lack the polished appearance of Grad-CAM outputs yet are prediction-critical under the
perturbation protocol: removing them collapses the model's confidence quickly. Under the native-readout hypothesis, this outcome is expected. Grad-CAM's smooth appearance is a \emph{consequence} of its structural distance from the CNN's decision mechanism --- the gradient-weighted activations are spatially low-pass and therefore visually pleasant --- while Rollout's sparsity is a \emph{consequence} of its proximity to the Transformer's attention-derived rollout maps, which are empirically concentrated in the sense that a small number of patches dominate the [CLS] readout for any given prediction. Thus, visual coherence is best treated as a qualitative diagnostic, while audit-facing explanation claims should be supported by perturbation-based faithfulness metrics.

\subsection{Cross-dataset analysis: boundary conditions}

The MVTec AD results (\S{}4.7) reveal that the native-readout hypothesis does not transfer straightforwardly to all settings. Three contrasts with WM-811K require explanation.

\textbf{Why does RISE compress on WM-811K but fail on MVTec?} On WM-811K, all four families train from scratch on simple geometric patterns (64$\times$64 grayscale). The resulting internal representations are comparably structured: each model learns to detect spatial clusters of defect pixels, and RISE's random masking effectively probes this shared decision boundary. On MVTec AD, models use different pretrained backbones (CNN vs. Transformer) that may encode substantially different feature hierarchies from ImageNet. RISE probes the model's actual input--output function, which may differ substantially across architectures in this pretrained setting; this offers one plausible explanation for the lack of compression. The mask-resolution ablation (\S{}4.8) argues against a resolution artefact as the explanation.

\textbf{Why is the native-method ranking reversed?} On WM-811K, ViT-Tiny's Attention Rollout achieves the best Deletion AUC despite weaker classification performance. On MVTec, however, ResNet18+CBAM's Grad-CAM is best, while ViT-Tiny's Rollout is weakest among native methods. This reversal is therefore unlikely to be explained by accuracy alone. A more plausible explanation is that ImageNet-pretrained spatial-hierarchy backbones are better matched to MVTec's texture-like anomaly patterns and binary defect-detection setting, whereas the global-attention ViT readout is less well aligned with the fine-tuned decision boundary. ViT-Tiny's lower MVTec accuracy (87 \% balanced accuracy vs. 92--94 \% for the three spatial-hierarchy models) may amplify this effect, but is better treated as an accompanying factor rather than the dominant cause.

\textbf{Why does ViT have the highest ground-truth IoU despite the worst Deletion AUC?} This dissociation between faithfulness-to-model (Deletion/Insertion) and faithfulness-to-ground-truth (IoU) is informative. Deletion AUC measures whether the heatmap identifies pixels the \emph{model} relies on; IoU measures whether the heatmap overlaps with the \emph{actual defect}. ViT's attention maps spread across the defect region (high IoU) but do not precisely rank the pixels the model uses for its decision (high Deletion AUC). This suggests that on MVTec, ViT's attention is spatially aware but not decision-precise --- consistent with its lower classification accuracy. This result cautions against treating localisation accuracy (IoU) and model- faithfulness (Deletion AUC) as interchangeable objectives.

Taken together, these results suggest that the native-readout hypothesis is most clearly supported when models share comparable internal representations, as in from-scratch training on domain-specific data, and when the explanation method's structural distance from the decision mechanism is the dominant source of variation. When pretrained models bring substantially different feature hierarchies, the architecture's representations themselves become the dominant factor, and the explainer pathway is secondary. The inclusion of Swin-Tiny sharpens this synthesis: on MVTec, all three spatial-hierarchy models (ResNet, DenseNet, Swin) have native Grad-CAM that outperforms RISE, while only the global-attention model (ViT) reverses this pattern. This holds regardless of whether the spatial hierarchy is implemented via convolutions (ResNet, DenseNet) or windowed self-attention (Swin), supporting the interpretation that the critical architectural property is the presence of hierarchical spatial feature maps rather than the specific computational primitive. It should be noted that the MVTec reversal admits multiple explanations --- including the classification-accuracy gap (ViT 87 \% vs others 92--94 \%), pretraining domain mismatch, and task-structure differences --- and the current data cannot uniquely distinguish among them.

The final-layer CLS attention ablation further clarifies the ViT failure mode on MVTec AD: final-layer attention (Del 0.857) performs worse than full Rollout (0.770), indicating that multi-layer integration remains beneficial. However, both native attention methods are outperformed by RISE (0.656), suggesting that the main limitation is not rollout depth alone. In the pretrained MVTec setting, native attention matrices appear only partially aligned with the fine-tuned decision boundary; RISE bypasses this pathway by directly probing the model's input--output response. This contrasts with WM-811K, where from-scratch training appears to produce attention matrices better aligned with the task-specific decision evidence.

These results delimit the scope of the hypothesis and clarify when explainer-pathway effects are likely to dominate representation-level differences.

\subsection{Practical recommendations}

Four deployment-oriented recommendations follow from the hypothesis tested above. First, when selecting an explainer, practitioners should inspect the model's \textbf{readout structure} rather than its architecture family. If the model produces hierarchical spatial feature maps (CNN or Swin-like architectures), Grad-CAM on the final spatial stage is the appropriate native method. If the model uses global attention with a [CLS] token (ViT-like architectures), Attention Rollout is preferred. This decision rule replaces the simpler but incorrect heuristic "CNN $\rightarrow$ Grad-CAM, Transformer $\rightarrow$ Rollout." Second, when explanations are presented to human operators for defect review, consistency across similar wafers is typically more important than peak median performance; ViT-Tiny's tight stability distribution (standard deviation 0.107) and Swin-Tiny's high stability (0.097)
therefore make them informative candidates for further investigation in operator-facing explanation displays when explanation consistency is prioritised, although their lower classification accuracy would need to be addressed before deployment. Third, when explanations feed downstream automation --- for example, masking a suspicious region for re-inspection --- Deletion and Insertion AUCs are the most relevant criteria; under the WM-811K zero-fill protocol, ViT-Tiny is the strongest native-method candidate, while RISE provides a stronger but more expensive offline audit reference. Fourth, when classification accuracy is the primary objective and interpretability is secondary, DenseNet121 or ResNet18+CBAM remain appropriate choices, \emph{provided} that the accompanying faithfulness metrics are reported alongside classification metrics to prevent over-claiming about what the heatmaps mean. At a design level, the hypothesis also suggests a forward-looking principle: architectures intended for
interpretable deployment should be chosen with their explanation pathway in mind from the outset, rather than retrofitted with a post-hoc explainer whose structural distance from the decision mechanism may limit measured faithfulness under a given protocol. In settings where classification accuracy is non-negotiable, the native-readout principle motivates future work on hybrid architectures that combine convolutional classification strength with attention-based explanation pathways --- for example, a CNN backbone with a lightweight Transformer head whose attention matrices serve as the explanation readout.

Table 13 summarises recommended audit choices by deployment objective, based on the evidence from both datasets.

\begin{center}
\small \textbf{Table 13.} Recommended explainer audit choices by deployment objective
\end{center}

\begin{tabular}{lll}
\toprule
Deployment objective & Recommended audit choice & Rationale \\
\midrule
Fast qualitative review & Native explainer & Low overhead, compatible \\
Offline faithfulness audit & RISE & Probes input--output directly \\
Wafer-map ViT explanation & Attention Rollout & Best WM-811K Del AUC \\
Diagnosing ViT failure & Final-layer CLS vs Rollout & Separates access from depth \\
CNN Grad-CAM reliability & Random baseline + RISE & Detects unfaithful priors \\
Pretrained-model inspection & Validate with RISE & Native may be misaligned \\
\bottomrule
\end{tabular}

\clearpage
\subsection{Practical audit workflow for industrial visual inspection}

Based on the WM-811K and MVTec AD results, we recommend treating explanation generation as an auditable component of industrial visual inspection systems rather than as a visualization-only post-processing step. Figure 8 summarises the proposed workflow.

The workflow is organized as an audit sequence rather than a visualization pipeline. Step [1] fixes the intended use of the explanation, since model audit, defect localization, human review, and downstream automation emphasize different criteria. Steps [2]--[3] establish predictive performance before interpretability is assessed. Step [4] selects the native explanation path from the model's readout structure: spatial-hierarchy models follow the Grad-CAM path [4a], whereas global CLS-token ViTs follow the Attention Rollout path [4b]. Steps [5]--[6] evaluate the resulting heatmaps using perturbation metrics and controls, including random baselines, RISE, and perturbation-fill sensitivity. If the explanation claim does not survive these controls, Step [7a] diagnoses the failure mode before Step [8] selects a qualified deployment strategy.

\begin{figure}[htbp]
\centering
\resizebox{0.9\linewidth}{!}{%
\begin{tikzpicture}[
    font=\sffamily\small,
    node distance=9mm and 13mm,
    hand/.style={
        decorate,
        decoration={random steps,segment length=4pt,amplitude=0.25pt}
    },
    arrow/.style={
        -{Stealth[round,length=5pt]},
        draw=black!55,
        line width=0.55pt,
        hand
    },
    box/.style={
        rectangle,
        rounded corners=3pt,
        draw=black!65,
        fill=black!3,
        line width=0.75pt,
        align=center,
        text width=33mm,
        minimum height=8mm,
        inner sep=4pt
    },
    keybox/.style={
        rectangle,
        rounded corners=5pt,
        draw=blue!55!black,
        fill=blue!4,
        line width=0.75pt,
        align=center,
        text width=33mm,
        minimum height=8mm,
        inner sep=4pt
    },
    decision/.style={
        diamond,
        aspect=1.8,
        draw=black!65,
        fill=black!2,
        line width=0.65pt,
        align=center,
        text width=28mm,
        inner sep=2pt
    },
    finalbox/.style={
        rectangle,
        rounded corners=5pt,
        draw=black!75,
        fill=black!5,
        line width=0.75pt,
        align=center,
        text width=35mm,
        minimum height=9mm,
        inner sep=4pt
    }
]

\node[keybox] (use) {[1] Define use};
\node[box, below=of use, xshift=0mm] (train) {[2] Train models};
\node[box, below=of train, xshift=0mm] (perf) {[3] Evaluate performance};

\node[decision, below=10mm of perf] (readout) {[4] Readout\\structure?};

\node[keybox, below left=10mm and 14mm of readout] (spatial) {[4a] Spatial\\hierarchy};
\node[keybox, below right=10mm and 14mm of readout] (global) {[4b] Global\\CLS attention};

\node[box, below=18mm of readout] (native) {[5] Native\\explanation};
\node[box, below=of native, xshift=0mm] (audit) {[6] Audit\\with controls};

\node[decision, below=10mm of audit] (survive) {[7] Claim\\survives?};

\node[box, below left=10mm and 12mm of survive] (diagnose) {[7a] Diagnose\\failure mode};
\node[finalbox, below right=10mm and 12mm of survive] (select) {[8] Select\\strategy};

\draw[arrow] (use) -- (train);
\draw[arrow] (train) -- (perf);
\draw[arrow] (perf) -- (readout);

\draw[arrow] (readout) -- node[above left, font=\scriptsize, text=black!60] {Grad-CAM path} (spatial);
\draw[arrow] (readout) -- node[above right, font=\scriptsize, text=black!60] {Rollout path} (global);

\draw[arrow] (spatial.south) |- (native.west);
\draw[arrow] (global.south) |- (native.east);

\draw[arrow] (native) -- (audit);
\draw[arrow] (audit) -- (survive);

\draw[arrow] (survive) -- node[above left, font=\scriptsize, text=black!60] {no / mixed} (diagnose);
\draw[arrow] (survive) -- node[above right, font=\scriptsize, text=black!60] {yes / qualified} (select);
\draw[arrow] (diagnose.east) |- (select.west);

\end{tikzpicture}
}
\caption{\textbf{Figure 8.} Decision-oriented architecture-aware explanation audit workflow. Step labels are expanded in the surrounding text.}
\end{figure}
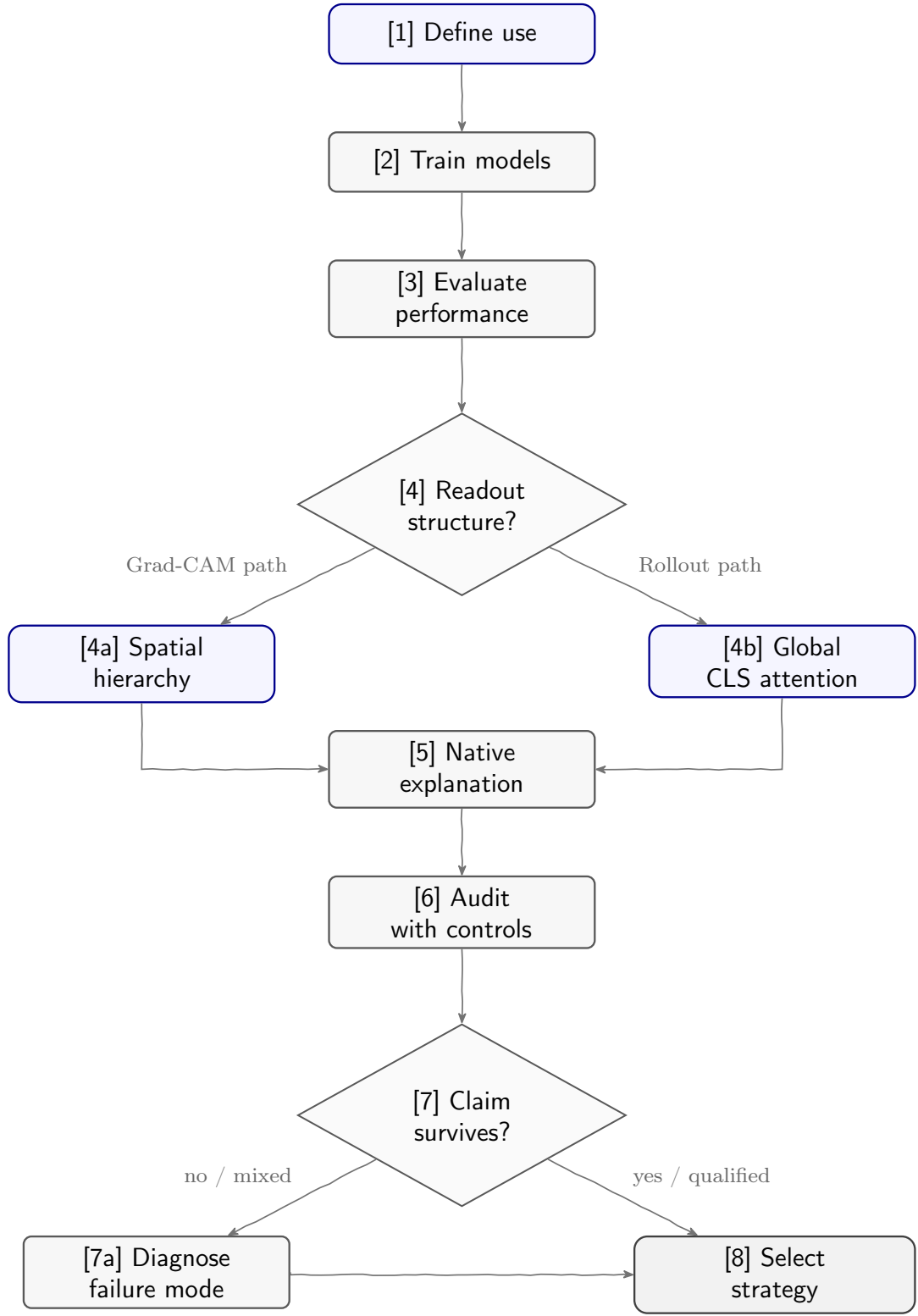

\subsection{Limitations and threats to validity}

Several limitations qualify the generalisability of the findings. The structural factors discussed in this paper are not fully orthogonal: readout directness, spatial granularity, and rollout path depth partly co-vary across the model--explainer pairs. In \S{}1.2, the native-readout hypothesis was operationalised through two primary proxies, readout directness and spatial granularity. The later ViT-specific ablations add a third factor, rollout path depth, because full Attention Rollout differs from final-layer CLS attention not only in readout origin but also in the number of layers through which information flow is accumulated. Two ablations partially disentangle these factors. Grad-CAM applied to ViT-Tiny (Supplementary Table S7) perturbs readout directness while holding the model and patch-level input granularity fixed, yielding Del 0.492, indistinguishable from random. Final-layer CLS attention (Supplementary Table S10) holds the readout origin fixed as native attention while removing multi-layer rollout depth, yielding Del 0.367. Together these ablations suggest that both direct access to native attention and multi-layer rollout contribute materially to the Rollout advantage. However, the decomposition remains approximate because the manipulations are not perfectly orthogonal: final-layer attention also changes how inter-layer information mixing is represented, and neither ablation fully isolates spatial granularity as an independent causal factor. The reported $\approx$42 \% / $\approx$52 \% partition should therefore be interpreted as a path-dependent decomposition of the observed WM-811K gap, not as a universal causal attribution.

The remaining limitations concern evaluation protocol and dataset coverage. The perturbation baseline used to compute Deletion and Insertion AUCs is zero-fill, which is semantically reasonable on WM-811K because the value zero encodes the wafer background, but which will not generalise to natural-image explainability tasks where blur- or mean-fill baselines are more conventional. Moreover, removing interior die pixels by setting them to zero can create out-of- distribution wafer patterns; the reported AUC values should therefore be interpreted as relative comparisons under a fixed perturbation operator rather than absolute faithfulness scores. Different baselines may shift the absolute AUC values and can affect the family ordering, as demonstrated by the blur-fill analysis in \S{}4.6. The 64$\times$64 input resolution constrains the granularity at which the metrics can reward
fine-grained attribution; re-running the experiment at 128$\times$128 would partially control for the spatial-granularity proxy. Each family was evaluated with a single representative interpretability method, and extending the comparison to multiple methods within each family --- Grad-CAM++ or Score-CAM for CNNs, the class-specific LRP+attention hybrid of Chefer et al. [10] for Transformers, plus model-agnostic methods such as LIME and Integrated Gradients as used by Khatun et al. [13] and Lee et al. [14] --- would place additional data points along the readout- directness proxy and provide a sharper test. Pairing faithfulness with probability-calibration diagnostics such as temperature scaling, as demonstrated for wafer maps by Lee et al. [14], would yield a more complete picture of deployment-grade explanation quality. Finally, the heavy dominance of the "None" class in WM-811K is only
partially compensated by stratified sampling; conclusions are therefore most robust for the defect classes rather than for the majority class. A sensitivity analysis over top-k ratios (5 \%, 10 \%, 20 \%) shows that the family ordering is robust to perturbation granularity (Supplementary Table S12). Because the bootstrap confidence intervals operate over pooled per-sample measurements across three seeds, they should be interpreted as uncertainty over the evaluated sample pool rather than as fully independent dataset-level uncertainty.

Two alternative accounts of the observed effect deserve explicit acknowledgment. First, because ViT-Tiny achieves a lower macro-F1 (0.747) than either CNN family (0.849 and 0.853), it is possible that a weaker classifier with a simpler decision boundary naturally relies on fewer, more spatially localised features, producing sparser explanations that score well on a metric that rewards sparse localisation --- independent of whether the explanation method is structurally privileged. Supplementary Table S11 addresses this directly: restricting evaluation to the 140 samples correctly classified by all four families preserves the full ranking (ViT Del 0.216, Swin 0.45, CNN 0.50--0.55), ruling out accuracy-driven confounding. Second, the zero-fill deletion protocol on WM-811K rewards methods that localise defect pixels, and patch-level attention does this by construction at the 4$\times$4 grid
used here; however, applying Grad-CAM to the \emph{same} ViT model (Supplementary Table S7) yields Deletion AUC 0.492 --- indistinguishable from random --- demonstrating that the WM-811K faithfulness advantage is attributable to the native-readout explainer, not to the ViT architecture or its patch granularity. The RISE experiment (\S{}4.5) provides the most direct adjudication: RISE applied uniformly to all four families compresses Deletion AUC to 0.091--0.130, indicating that the gap is attributable to the explainer pathway. Had RISE preserved the ViT advantage, the effect would have been attributable to the architecture's representations; the observed compression argues against that alternative account under the WM-811K perturbation protocol.

The MVTec AD boundary-condition study (\S{}4.7) introduces additional limitations specific to the second dataset. ViT-Tiny's lower classification accuracy (87 \% balanced accuracy vs. 92--94 \% for the three spatial-hierarchy models) is a confound that cannot be fully resolved without a ViT variant matching CNN accuracy on this task. The use of different pretrained backbones (ImageNet CNN vs. ImageNet ViT) means the comparison tests the joint effect of architecture + pretraining rather than architecture alone. MVTec interpretability is evaluated on seed 42 only (unlike WM-811K's three-seed protocol), so between-run variance on this dataset is not characterised. RISE performs worse than native methods for the two CNN families on MVTec (Del 0.886--0.938 vs native 0.413--0.671) and only modestly improves ViT-Tiny (Del 0.656 vs native 0.770), failing to produce the WM-811K-style compression. The mask-resolution ablation
(\S{}4.8) rules out one simple explanation but does not establish that RISE was globally optimized for MVTec AD (other hyperparameters such as mask count, probability, and fill value remain untested).

The native-readout hypothesis is supported by the WM-811K evidence and the RISE control, with boundary conditions identified on MVTec AD. Two datasets with four model families, two explanation paradigms (native + model-agnostic), and a mask-resolution ablation provide converging evidence, but do not establish a universal principle. Additional ablations that would further tighten the conclusions include: a ViT with larger patch size (P=8, matching CNN spatial resolution, to directly test the granularity proxy), and multi-seed interpretability evaluation on MVTec AD. The blur-fill perturbation-baseline analysis (\S{}4.6) further qualifies the WM-811K conclusions as protocol-specific and motivates reporting the perturbation baseline alongside faithfulness metrics in deployment audits. The contribution is to formulate the hypothesis precisely, support it under a specified controlled protocol, and identify the settings where it requires qualification.

\section{Conclusion}

This paper presented an architecture-aware audit protocol for evaluating explanation faithfulness in industrial visual inspection. Rather than treating explanation methods as interchangeable post-hoc visualization tools, we examined whether measured faithfulness under a documented perturbation protocol depends on the compatibility between a model's readout structure and the explainer used to interpret it.

On WM-811K, the results support the native-readout hypothesis under the zero-fill perturbation protocol: ViT-Tiny paired with Attention Rollout achieved the lowest (most faithful) native-explainer Deletion AUC, while CNN-based models were better matched with Grad-CAM. Swin-Tiny provided an important diagnostic case, showing that explanation compatibility is better understood in terms of readout structure than broad architecture family: although Transformer-based, its hierarchical spatial feature maps made Grad-CAM a structurally appropriate and competitive explainer. At the same time, RISE outperformed all native methods on WM-811K, demonstrating that native readout should be interpreted as a compatibility principle rather than a guarantee of optimality.

The additional sensitivity and boundary-condition analyses further qualify these findings. The blur-fill sensitivity analysis further showed that Deletion AUC rankings are not explainer-intrinsic, but depend on the perturbation operator included in the audit protocol. The exploratory MVTec AD study also showed that trends observed on wafer maps do not necessarily transfer directly to anomaly detection, where dataset structure, pretraining, binary classification, and localization objectives interact with explanation behavior.

This paper does not establish a universally valid ranking of explainers. Instead, it shows that explanation reliability in industrial visual inspection is conditional on the model architecture, explainer readout path, and perturbation operator. The practical lesson is that heatmaps should not be trusted solely because they are visually plausible, architecture-conventional, or even architecture-native. Explanation methods should be selected and validated through a documented, architecture-aware audit protocol; for deployment-facing systems, heatmaps should be accompanied by quantitative perturbation-faithfulness scores, model-agnostic audit baselines, and explicit reporting of the perturbation protocol. This reframes explanation generation from a visualization step into an auditable component of industrial AI deployment.

\subsection*{References}

\textbf{Foundational architectures and interpretability methods}

\begin{list}{}{\setlength{\leftmargin}{2.5em}\setlength{\labelwidth}{2em}\setlength{\labelsep}{0.3em}\setlength{\itemsep}{0.2ex}}
\item[1.] He, K., Zhang, X., Ren, S., \& Sun, J. \emph{Deep Residual Learning for Image Recognition}. CVPR, 2016. \url{https://www.cv-foundation.org/openaccess/content_cvpr_2016/papers/He_Deep_Residual_Learning_CVPR_2016_paper.pdf}
\item[2.] Woo, S., Park, J., Lee, J.-Y., \& Kweon, I. S. \emph{CBAM: Convolutional Block Attention Module}. ECCV, 2018. \url{https://arxiv.org/abs/1807.06521}
\item[3.] Huang, G., Liu, Z., van der Maaten, L., \& Weinberger, K. Q. \emph{Densely Connected Convolutional Networks}. CVPR, 2017. \url{https://openaccess.thecvf.com/content_cvpr_2017/papers/Huang_Densely_Connected_Convolutional_CVPR_2017_paper.pdf}
\item[4.] Dosovitskiy, A. et al. \emph{An Image Is Worth 16$\times$16 Words: Transformers for Image Recognition at Scale}. ICLR, 2021. \url{https://dblp.org/rec/conf/iclr/DosovitskiyB0WZ21}
\item[5.] Selvaraju, R. R. et al. \emph{Grad-CAM: Visual Explanations from Deep Networks via Gradient-Based Localization}. ICCV, 2017. \url{https://openaccess.thecvf.com/content_ICCV_2017/papers/Selvaraju_Grad-CAM_Visual_Explanations_ICCV_2017_paper.pdf}
\item[6.] Abnar, S., \& Zuidema, W. \emph{Quantifying Attention Flow in Transformers}. ACL, 2020. \url{https://arxiv.org/abs/2005.00928}
\item[7.] Petsiuk, V., Das, A., \& Saenko, K. \emph{RISE: Randomized Input Sampling for Explanation of Black-Box Models}. BMVC, 2018. \url{https://arxiv.org/abs/1806.07421}
\item[8.] Alvarez-Melis, D., \& Jaakkola, T. S. \emph{On the Robustness of Interpretability Methods}. ICML Workshop, 2018. \url{https://arxiv.org/abs/1806.08049}
\item[9.] Jain, S., \& Wallace, B. C. \emph{Attention Is Not Explanation}. NAACL, 2019. \url{https://doi.org/10.48550/arXiv.1902.10186}
\item[10.] Chefer, H., Gur, S., \& Wolf, L. \emph{Transformer Interpretability Beyond Attention Visualization}. CVPR, 2021. \url{https://www.computer.org/csdl/proceedings-article/cvpr/2021/450900a782/1yeIsbbCMO4}
\item[11.] Sundararajan, M., Taly, A., \& Yan, Q. \emph{Axiomatic Attribution for Deep Networks} (Integrated Gradients). ICML, 2017. \url{https://arxiv.org/abs/1703.01365}
\item[12.] Zeiler, M. D., \& Fergus, R. \emph{Visualizing and Understanding Convolutional Networks} (Occlusion sensitivity). ECCV, 2014. \url{https://cs.nyu.edu/~fergus/papers/zeilerECCV2014.pdf}
\end{list}

\textbf{Recent wafer-map defect classification and XAI (2024--2026)}

\begin{list}{}{\setlength{\leftmargin}{2.5em}\setlength{\labelwidth}{2em}\setlength{\labelsep}{0.3em}\setlength{\itemsep}{0.2ex}}
\item[13.] Khatun, M. R., Farid, F. A., Dhar, S., Islam, M. S., Uddin, J., \& Abdul Karim, H. \emph{CBAM-Enhanced Lightweight CNN for Wafer Map Defect Classification}. Frontiers in Electronics, 7:1750707, 2026. \url{https://www.frontiersin.org/journals/electronics/articles/10.3389/felec.2026.1750707/full}
\item[14.] Lee, J., Ju, Y., Lim, J., Hong, S., Baek, S.-W., \& Lee, J. \emph{Enhancing Confidence and Interpretability of a CNN-Based Wafer Defect Classification Model Using Temperature Scaling and LIME}. Micromachines, 16(9):1057, 2025. \url{https://www.mdpi.com/2072-666X/16/9/1057}
\item[15.] Lee, C.-Y., Pleva, M., Hl\'adek, D., Lee, C.-W., \& Su, M.-H. \emph{Ensemble Learning for Wafer Defect Pattern Classification in the Semiconductor Industry}. IEEE Access, 13, 2025. \url{https://ieeexplore.ieee.org/iel8/6287639/10820123/11145756.pdf}
\item[16.] Park, S. Y., \& Kim, T. S. \emph{Fuzzy Inference System for Interpretable Classification of Wafer Map Defect Patterns}. Electronics, 15(1):130, 2026. \url{https://www.mdpi.com/2079-9292/15/1/130}
\item[17.] Pilli, V. S. R. R. \emph{Intelligent Model to Detect and Classify Silicon Wafer Map Images}. M.Sc. Thesis, Purdue University, 2024. \url{https://hammer.purdue.edu/ndownloader/files/49412641}
\end{list}

\textbf{Benchmark dataset}

\begin{list}{}{\setlength{\leftmargin}{2.5em}\setlength{\labelwidth}{2em}\setlength{\labelsep}{0.3em}\setlength{\itemsep}{0.2ex}}
\item[18.] Wu, M.-J., Jang, J.-S. R., \& Chen, J.-L. \emph{Wafer Map Failure Pattern Recognition and Similarity Ranking for Large-Scale Data Sets}. IEEE Transactions on Semiconductor Manufacturing, 28(1):1--12, 2015. \url{https://ui.adsabs.harvard.edu/abs/2015ITSM...28S4237W/abstract}
\end{list}

\textbf{Boundary-condition dataset and pretrained models}

\begin{list}{}{\setlength{\leftmargin}{2.5em}\setlength{\labelwidth}{2em}\setlength{\labelsep}{0.3em}\setlength{\itemsep}{0.2ex}}
\item[19.] Bergmann, P., Fauser, M., Sattlegger, D., \& Steger, C. \emph{MVTec AD --- A Comprehensive Real-World Dataset for Unsupervised Anomaly Detection}. CVPR, 2019. \url{https://openaccess.thecvf.com/content_CVPR_2019/papers/Bergmann_MVTec_AD_--_A_Comprehensive_Real-World_Dataset_for_Unsupervised_Anomaly_CVPR_2019_paper.pdf}
\item[20.] Touvron, H., Cord, M., Douze, M., Massa, F., Sablayrolles, A., \& J\'egou, H. \emph{Training Data-Efficient Image Transformers \& Distillation Through Attention} (DeiT). ICML, 2021. \url{https://arxiv.org/abs/2012.12877}
\end{list}

\textbf{Trustworthy industrial AI}

\begin{list}{}{\setlength{\leftmargin}{2.5em}\setlength{\labelwidth}{2em}\setlength{\labelsep}{0.3em}\setlength{\itemsep}{0.2ex}}
\item[21.] Breque, M., De Nul, L., \& Petridis, A. \emph{Industry 5.0: Towards a Sustainable, Human-Centric and Resilient European Industry}. European Commission, Directorate-General for Research and Innovation, 2021. \url{https://op.europa.eu/publication/manifestation_identifier/PUB_KIBD20021ENN}
\item[22.] Moosavi, S., Farajzadeh-Zanjani, M., Razavi-Far, R., Palade, V., \& Saif, M. \emph{Explainable AI in Manufacturing and Industrial Cyber--Physical Systems: A Survey}. Electronics, 13(17):3497, 2024. \url{https://www.mdpi.com/2079-9292/13/17/3497}
\end{list}

\textbf{Hierarchical vision transformers}

\begin{list}{}{\setlength{\leftmargin}{2.5em}\setlength{\labelwidth}{2em}\setlength{\labelsep}{0.3em}\setlength{\itemsep}{0.2ex}}
\item[23.] Liu, Z., Lin, Y., Cao, Y., Hu, H., Wei, Y., Zhang, Z., Lin, S., \& Guo, B. \emph{Swin Transformer: Hierarchical Vision Transformer using Shifted Windows}. ICCV, 2021. \url{https://arxiv.org/abs/2103.14030}
\end{list}

\clearpage
\appendix
\phantomsection
\section*{Supplementary Material}
\addcontentsline{toc}{section}{Supplementary Material}

\phantomsection
\subsection*{S1. Experimental workflow}
\addcontentsline{toc}{subsection}{S1. Experimental workflow}

The full pipeline proceeds in four stages, from raw data ingestion through to the final report. Each stage is automated by dedicated scripts, enabling full reproduction with a single command (\texttt{bash rerun\_report\_journal.sh} for WM-811K; \texttt{bash run\_mvtec\_parallel.sh} + \texttt{python interpret\_eval\_mvtec.py} for MVTec AD).

\begin{Verbatim}
+--------------+     +------------------+     +----------------------+
|  LSWMD.pkl   |---->|  Preprocessing   |---->|  Lot-Group Split     |
|  (811K maps) |     |  64x64, nan->0    |     |  70 / 15 / 15        |
+--------------+     +------------------+     +----------+-----------+
                                                         |
+--------------+     +------------------+     +----------v-----------+
|  MVTec AD    |---->|  Resize 256x256  |---->|  Stratified Split    |
|  (5354 imgs) |     |  ImageNet norm   |     |  70 / 15 / 15        |
+--------------+     +------------------+     +----------+-----------+
                                                         |
                     +-----------------------------------+|
                     |      TRAINING  (x12 WM + x4 MVTec)||
                     |  4 families x 3 seeds (WM-811K)   |<+
                     |  4 families x 1 seed  (MVTec AD)  |
                     |  AdamW . cosine LR . AMP          |
                     |  Early stop: val macro-F1         |
                     +---------------+-------------------+
                                     |
                     +---------------v-------------------+
                     |      CLASSIFICATION EVAL           |
                     |  Acc, Bal-Acc, F1-Macro, MCC       |
                     +---------------+-------------------+
                                     |
                     +---------------v-------------------+
                     |     INTERPRETABILITY EVAL          |
                     |  Native methods + RISE             |
                     |  Del AUC . Ins AUC . Stability    |
                     |  + IoU (MVTec AD only)             |
                     +---------------+-------------------+
                                     |
                     +---------------v-------------------+
                     |        REPORT GENERATION           |
                     |  Report -> HTML + saved figures     |
                     +-----------------------------------+
\end{Verbatim}

\begin{center}
\small \textbf{Supplementary Diagram S1.} End-to-end experimental pipeline (both datasets)
\end{center}

\clearpage
\phantomsection
\subsection*{S2. Code overview}
\addcontentsline{toc}{subsection}{S2. Code overview}

\begin{Verbatim}
+-----------------------------------------------------------------+
|                       CLI ENTRY POINTS                           |
+----------+--------------+------------------+--------------------+
| main.py  | compare.py   | interpret_eval.py| generate_panel.py  |
|main_mvtec|              |interpret_eval_   | rise_eval.py       |
|          |              |  mvtec.py        |                    |
+----+-----+------+-------+-------+----------+--------------------+
     |            |               |
     v            v               v
+-----------------------------------------------------------------+
|                          src/                                    |
+------------+------------+--------------+------------------------+
| dataset.py | models.py  |  train.py    | interpretability.py    |
|dataset_    |models_     |  AMP/LR/Ckpt| Heatmap/Del/Ins/Stab  |
|  mvtec.py  |  mvtec.py  |             |                        |
+------------+------------+--------------+------------------------+
     |                                          |
     v                                          v
+-----------------------------------------------------------------+
|                         utils/                                   |
+------------------+------------------+---------------------------+
| metrics.py       | visualize.py     | augmentation.py           |
| FocalLoss/Eval   | GradCAM/Rollout  | Rotation/flip/noise       |
+------------------+------------------+---------------------------+
\end{Verbatim}

\begin{center}
\small \textbf{Supplementary Diagram S2.} Module dependency graph. MVTec-specific modules (suffixed \_mvtec) mirror the WM-811K modules with RGB/binary adaptations
\end{center}

The codebase embodies four design principles. First, every experiment is \textbf{config-driven}: all hyperparameters reside in YAML files (\texttt{configs/} for WM-811K, \texttt{configs\_mvtec/} for MVTec AD), eliminating magic numbers from source code and allowing exact reconstruction of any run from its saved configuration. Second, a \textbf{registry pattern} in \texttt{src/models.py} and \texttt{src/models\_mvtec.py} maps string names to model classes. Third, a \textbf{lot-group split} at the data layer prevents the information leakage that would otherwise arise from same-lot wafers sharing systematic defects (WM-811K); MVTec AD uses stratified random splitting. Finally, the project maintains \textbf{strict separation of concerns}: training, evaluation, and interpretability live in independent modules, with dataset-specific variants kept in parallel files rather than conditional
branches.

\phantomsection
\subsection*{S3. Training curves}
\addcontentsline{toc}{subsection}{S3. Training curves}

Per-family training and validation loss curves (seed 42) are shown in Supplementary Figure S1.

\begin{figure}[htbp]
\centering
\includegraphics[width=0.70\linewidth]{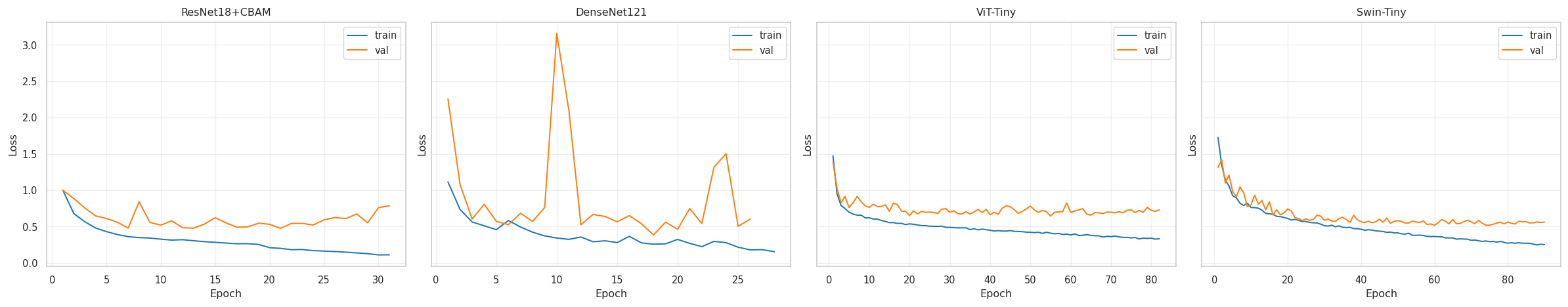}
\caption{\textbf{Supplementary Figure S1.} Training and validation loss curves, WM-811K (seed 42)}
\end{figure}

\phantomsection
\subsection*{S4. Complete hyperparameter tables}
\addcontentsline{toc}{subsection}{S4. Complete hyperparameter tables}

\begin{table}[htbp]
\centering
\small
\caption{\textbf{Supplementary Table S1a.} Training hyperparameters per family (WM-811K)}
\resizebox{\linewidth}{!}{%
\begin{tabular}{llrrrrlrllrlr}
\toprule
Family & Optimizer & Learning rate & Weight decay & Batch size & Max epochs & Scheduler & Early stop patience & Early stop metric & Loss & Grad clip norm & Mixed precision & Dropout \\
\midrule
ResNet18+CBAM & adam & 0.0010 & 0.00 & 32 & 50 & reduce\_on\_plateau & 10 & val\_balanced\_acc & cross\_entropy & 1.0 & True & 0.5 \\
DenseNet121 & adam & 0.0010 & 0.00 & 32 & 50 & reduce\_on\_plateau & 10 & val\_balanced\_acc & cross\_entropy & 1.0 & True & 0.5 \\
ViT-Tiny & adamw & 0.0003 & 0.05 & 64 & 100 & cosine & 20 & val\_f1\_macro & cross\_entropy & 1.0 & True & 0.3 \\
Swin-Tiny & adamw & 0.0003 & 0.05 & 64 & 100 & cosine & 20 & val\_f1\_macro & cross\_entropy & 1.0 & True & 0.3 \\
\bottomrule
\end{tabular}
}
\end{table}

All four WM-811K families share identical data and augmentation settings: 64$\times$64 images, lot-group split (70/15/15), cached preprocessing, augmentation with rotation probability 0.7, flip probability 0.5, Gaussian noise ($\sigma$=0.05, probability 0.3), and minority-class boosting enabled.

\begin{center}
\small \textbf{Supplementary Table S1b.} Training hyperparameters per family (MVTec AD)
\end{center}

\begin{tabular}{lcccc}
\toprule
Parameter & ResNet18+CBAM & DenseNet121 & ViT-Tiny (DeiT) & Swin-Tiny \\
\midrule
Pretrained & ImageNet & ImageNet & ImageNet (DeiT) & ImageNet \\
Optimizer & AdamW (all) &  &  &  \\
Learning rate & 0.0003 & 0.0003 & 0.0001 & 0.0001 \\
Weight decay & 0.05 (all) &  &  &  \\
Batch / epochs & 32 / 100 (all) &  &  &  \\
Scheduler & cosine (all) &  &  &  \\
Early stop & 15 / val\_f1 (all) &  &  &  \\
Loss / AMP / Drop & CE / $\checkmark$ / 0.5 (all) &  &  &  \\
Image / window & 256\textsuperscript{2} / --- & 256\textsuperscript{2} / --- & 256\textsuperscript{2} / --- & 256\textsuperscript{2} / 7 \\
Split & strat. 70/15/15 (all) &  &  &  \\
\bottomrule
\end{tabular}

\begin{center}
\small \textbf{Supplementary Table S2.} Interpretability evaluation parameters (both datasets)
\end{center}

\begin{tabular}{ll}
\toprule
Parameter & Value \\
\midrule
Samples per eval & WM-811K: 198/seed (9 classes); MVTec: 200 \\
Top-k ratio & 10 \% \\
Deletion/Insertion steps & 20 \\
Perturbation baseline & Zero-fill \\
Stability augmentations & 5 per sample \\
Stability transforms & Rot $\pm$15\textdegree{}, trans $\pm$3 px, noise $\sigma$=0.02 \\
Stability metric & Cosine similarity \\
Seeds & WM-811K: 7/123/456 (Swin), 42/123/456 (others); MVTec: 42 \\
RISE masks (N) & 4,000 \\
RISE mask resolution & 8 (default); ablation: 16, 32 (MVTec) \\
RISE probability (p) & 0.5 \\
IoU threshold & MVTec: binarised at 50th percentile \\
\bottomrule
\end{tabular}

\begin{table}[htbp]
\centering
\small
\caption{\textbf{Supplementary Table S3.} Per-seed classification results for all twelve WM-811K paper runs}
\resizebox{\linewidth}{!}{%
\begin{tabular}{lrrrrr}
\toprule
Family & Seed & Accuracy & F1 Macro & Balanced Acc & Epochs \\
\midrule
ResNet18+CBAM & 42 & 0.952 & 0.843 & 0.901 & 31 \\
ResNet18+CBAM & 123 & 0.960 & 0.850 & 0.902 & 35 \\
ResNet18+CBAM & 456 & 0.959 & 0.855 & 0.899 & 36 \\
DenseNet121 & 42 & 0.959 & 0.844 & 0.912 & 28 \\
DenseNet121 & 123 & 0.960 & 0.846 & 0.897 & 36 \\
DenseNet121 & 456 & 0.967 & 0.868 & 0.895 & 35 \\
ViT-Tiny & 42 & 0.920 & 0.744 & 0.819 & 82 \\
ViT-Tiny & 123 & 0.927 & 0.739 & 0.792 & 67 \\
ViT-Tiny & 456 & 0.938 & 0.758 & 0.786 & 94 \\
Swin-Tiny & 7 & 0.955 & 0.791 & 0.809 & 90 \\
Swin-Tiny & 123 & 0.953 & 0.787 & 0.814 & 78 \\
Swin-Tiny & 456 & 0.951 & 0.797 & 0.842 & 97 \\
\bottomrule
\end{tabular}
}
\end{table}

\phantomsection
\subsection*{S5. Raw qualitative heatmaps (no wafer overlay)}
\addcontentsline{toc}{subsection}{S5. Raw qualitative heatmaps (no wafer overlay)}

Figure 5 in the main text displays each heatmap overlaid on the input wafer for readability. That composite can visually blur the intrinsic behaviour of the heatmap with the wafer background. Supplementary Figure S2 shows the same samples with the wafer stripped away and each heatmap normalised to its own dynamic range. The raw view makes it clear that (i) ResNet18+CBAM's Grad-CAM is essentially input-independent --- a near-identical upper-left blob across all nine classes; (ii) DenseNet121's Grad-CAM is similarly quadrant-locked, with the quadrant chosen by class but no shape awareness within the quadrant; and (iii) ViT-Tiny's Attention Rollout genuinely adapts to the input, with clearest spatial correspondence on Center and Loc (hot-spots on the defect cluster) and Edge-Ring (hot-spots distributed around the ring circumference); alignment is weaker on Donut, Edge-Loc, and Near-Full.
For None the heatmap lacks a coherent defect-shaped structure (as expected given the absence of defect signal); for Random the activation pattern is distinguishable from None but does not track the diffuse defect pixels; and for Scratch the hot-spots partially follow the arc-shaped trajectory --- consistent with the Spearman rank correlations reported in Supplementary Table S4.

\begin{figure}[htbp]
\centering
\includegraphics[width=0.60\linewidth]{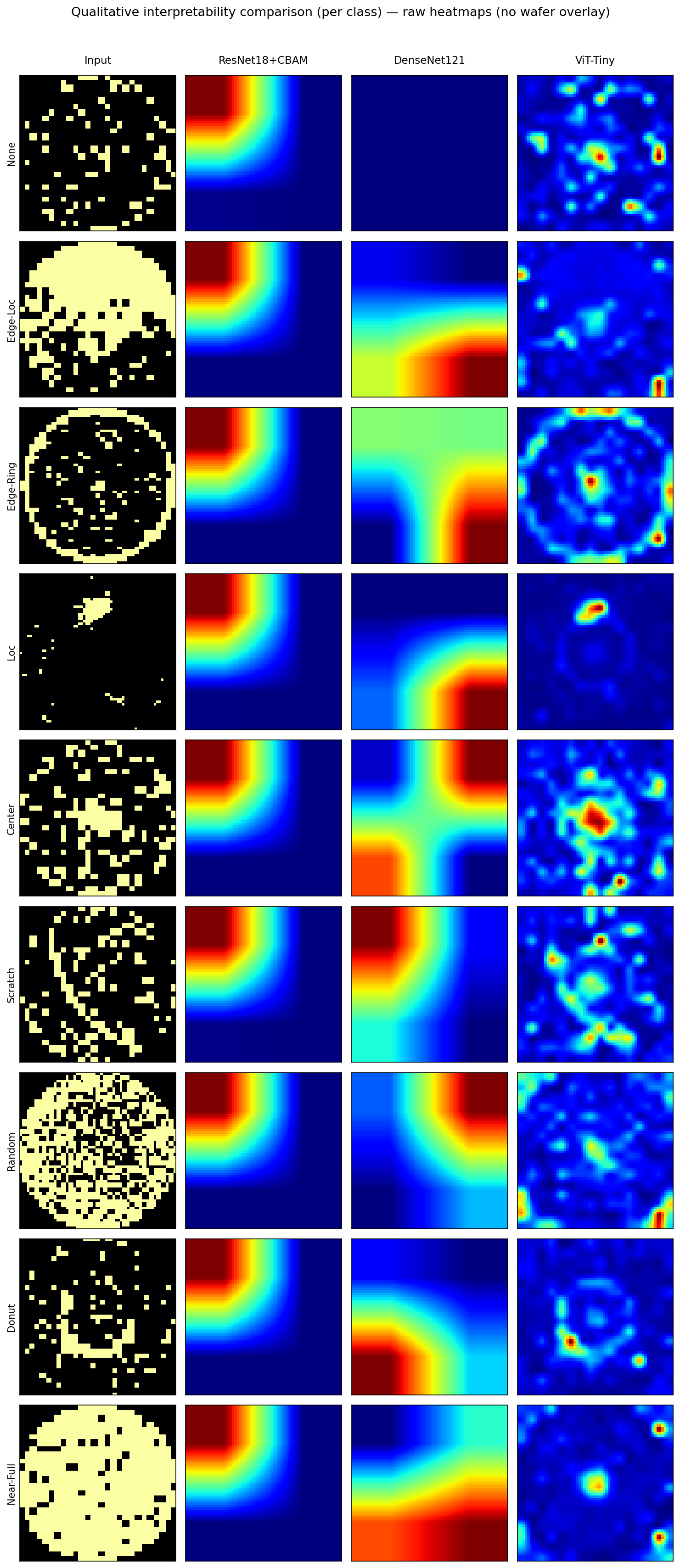}
\caption{\textbf{Supplementary Figure S2.} Raw qualitative heatmaps, WM-811K (no wafer overlay). Same samples as Figure 5; each heatmap is normalised to its own min--max range for a fair visual comparison}
\end{figure}

\clearpage
\phantomsection
\subsection*{S6. Spatial alignment with defect distribution}
\addcontentsline{toc}{subsection}{S6. Spatial alignment with defect distribution}

To quantify how well each method's heatmap aligns with the wafer's defect-pixel distribution, we compute the Spearman rank correlation between the heatmap and the binary defect mask, per sample, across all 594 evaluation samples per family. A correlation near zero indicates that the heatmap is not spatially aligned with the defects; a high positive correlation indicates that the heatmap concentrates intensity on defect pixels. This correlation is not used as a causal faithfulness metric; it only measures spatial alignment with labelled fail-die pixels. Supplementary Table S4 reports the mean and standard deviation per family.

\begin{table}[htbp]
\centering
\small
\caption{\textbf{Supplementary Table S4.} Spearman rank correlation between heatmap intensity and wafer defect-pixel distribution, WM-811K (n=594 samples per family). Values near zero indicate no spatial alignment}
\begin{tabular}{lrr}
\toprule
Family & Defect correlation (mean) & Defect correlation (std) \\
\midrule
ResNet18+CBAM & 0.037 & 0.105 \\
DenseNet121 & 0.015 & 0.126 \\
ViT-Tiny & 0.318 & 0.209 \\
Swin-Tiny & 0.042 & 0.105 \\
\bottomrule
\end{tabular}
\end{table}

\phantomsection
\subsection*{S7. Per-seed Deletion AUC}
\addcontentsline{toc}{subsection}{S7. Per-seed Deletion AUC}

To support the claim in \S{}5.1 that the faithfulness ranking is stable across random seeds, the per-seed mean Deletion AUC is reported below. Each cell is averaged across 200 balanced-stratified test samples.

\begin{table}[htbp]
\centering
\small
\caption{\textbf{Supplementary Table S5.} Per-seed mean Deletion AUC, WM-811K (200 balanced-stratified samples per seed)}
\begin{tabular}{lllrr}
\toprule
Family & seed 7 & seed 42 & seed 123 & seed 456 \\
\midrule
DenseNet121 & NaN & 0.544 & 0.514 & 0.519 \\
ResNet18+CBAM & NaN & 0.486 & 0.512 & 0.485 \\
Swin-Tiny & 0.44 & NaN & 0.413 & 0.444 \\
ViT-Tiny & NaN & 0.221 & 0.223 & 0.188 \\
\bottomrule
\end{tabular}
\end{table}

\phantomsection
\subsection*{S8. Ablation: random-pixel baseline}
\addcontentsline{toc}{subsection}{S8. Ablation: random-pixel baseline}

To calibrate the Deletion/Insertion metrics, we replace each method's heatmap ranking with a uniformly random pixel permutation and re-run the same perturbation protocol on the seed-42 models (198 stratified samples).

\begin{table}[htbp]
\centering
\small
\caption{\textbf{Supplementary Table S6.} Heatmap-based vs random-pixel Deletion/Insertion AUC, WM-811K (seed 42)}
\begin{tabular}{lrrrr}
\toprule
family & Del (heatmap) & Ins (heatmap) & Del (random) & Ins (random) \\
\midrule
densenet121 & 0.544 & 0.501 & 0.508 & 0.494 \\
resnet18\_cbam & 0.486 & 0.531 & 0.498 & 0.492 \\
swin\_tiny & 0.440 & 0.448 & 0.475 & 0.453 \\
vit\_tiny & 0.221 & 0.689 & 0.480 & 0.480 \\
\bottomrule
\end{tabular}
\end{table}

CNN Grad-CAM rankings perform no better than random ordering (ResNet Del 0.486 vs random 0.498; DenseNet Del 0.544 vs random 0.508 --- both at or above the random level), indicating that the highlighted pixels carry negligible prediction-relevant signal beyond chance. ViT-Tiny Rollout (Del 0.221) is far below the random baseline (0.480), providing stronger perturbation-based evidence of faithfulness.

On MVTec AD, RISE Deletion AUC remains high (0.66--0.94 across families and mask resolutions), consistent with limited faithfulness under this perturbation protocol on the MVTec dataset (\S{}4.7).

\clearpage
\phantomsection
\subsection*{S9. Ablation: Grad-CAM on ViT (non-native control)}
\addcontentsline{toc}{subsection}{S9. Ablation: Grad-CAM on ViT (non-native control)}

To disentangle architecture effects from explainer effects, we apply Grad-CAM (targeting the last Transformer encoder layer) to the same ViT-Tiny model that Attention Rollout explains. This is a deliberately non-native control: Grad-CAM was designed for convolutional feature maps, not Transformer token sequences.

\begin{table}[htbp]
\centering
\small
\caption{\textbf{Supplementary Table S7.} ViT-Tiny: native (Rollout) vs non-native (Grad-CAM) explainer, WM-811K}
\begin{tabular}{lrr}
\toprule
Method & Deletion AUC $\downarrow$ & Insertion AUC $\uparrow$ \\
\midrule
Attention Rollout (native) & 0.221 & 0.689 \\
Grad-CAM (non-native) & 0.492 & 0.488 \\
\bottomrule
\end{tabular}
\end{table}

Grad-CAM on ViT produces faithfulness indistinguishable from random (Del 0.492 vs random 0.480), while Rollout on the same model achieves Del 0.221. This indicates that the faithfulness advantage is attributable to the native-readout explainer, not to the ViT architecture or its patch-level spatial granularity.

\phantomsection
\subsection*{S10. Ablation: Final-layer CLS attention (rollout-depth control)}
\addcontentsline{toc}{subsection}{S10. Ablation: Final-layer CLS attention (rollout-depth control)}

To disentangle readout directness from rollout depth, we evaluate final-layer CLS-to-patch attention on ViT-Tiny: the last transformer block's attention weights from the [CLS] query to all patch tokens, averaged over heads. This holds the readout origin constant (native attention matrices) while removing the 12-layer recursive propagation used by full Attention Rollout.

\begin{tabular}{llll}
\toprule
Method & Readout & Depth & Del AUC $\downarrow$ \\
\midrule
Full Rollout & Native attention & 12-layer recursive & 0.211 \\
Final-layer CLS attn & Native attention & Single layer & 0.367 \\
Grad-CAM on ViT (S9) & Non-native (gradient) & Single layer & 0.492 \\
CNN Grad-CAM & Non-native (gradient) & Single layer & 0.495--0.525 \\
\bottomrule
\end{tabular}

\textbf{Table S10.} 2$\times$2 decomposition of the ViT faithfulness advantage. Final-layer CLS attention (3-seed mean, n=198 per seed). Seeds: Del 0.340 / 0.401 / 0.359 (mean 0.367 $\pm$ 0.031).

Implementation note: final-layer attention is computed as the raw [CLS]-to-patch row of the last block's attention matrix (averaged over heads), without the residual correction (\textonehalf{}A + \textonehalf{}I) used in full Rollout. This isolates the single-layer attention signal.

The total gap between full Rollout and CNN Grad-CAM is $\approx$0.30. Final-layer attention falls at 0.367, partitioning the gap into:

\begin{itemize}
\item \textbf{Readout directness} (final-layer attn vs Grad-CAM-on-ViT): 0.492 $-$ 0.367 = 0.125 ($\approx$42\% of total gap)
\item \textbf{Rollout depth} (full Rollout vs final-layer attn): 0.367 $-$ 0.211 = 0.156 ($\approx$52\% of total gap)
\end{itemize}

Both structural proxies carry independent signal. Readout directness alone (single-layer native attention) already improves over the non-native Grad-CAM-on-ViT control, while multi-layer rollout contributes a comparable additional improvement ($\approx$0.16) by accumulating information flow across the full transformer depth. The partition is path-dependent because the two manipulations are not strictly additive (the residual $\approx$6 \% reflects non-additivity).

\textbf{MVTec AD comparison (seed 42, n=200 defective samples):}

\begin{tabular}{ll}
\toprule
Method & Del AUC $\downarrow$ \\
\midrule
RISE & 0.656 \\
Full Rollout & 0.770 \\
Final-layer CLS attn & 0.857 \\
\bottomrule
\end{tabular}

On MVTec, the pattern reverses: final-layer attention is worst, full Rollout partially improves, but RISE outperforms both native methods. This indicates that the MVTec boundary condition is not simply a rollout-depth problem; pretrained attention pathways appear only partially aligned with the fine-tuned decision boundary.

\phantomsection
\subsection*{S11. Ablation: commonly-correct subset}
\addcontentsline{toc}{subsection}{S11. Ablation: commonly-correct subset}

To rule out the alternative explanation that a weaker classifier produces artificially faithful explanations (because simpler decision boundaries rely on fewer pixels), we restrict evaluation to the 140 samples (of 198) that all four families classify correctly in their primary seed.

\begin{table}[htbp]
\centering
\small
\caption{\textbf{Supplementary Table S11.} Faithfulness on commonly-correct samples only, WM-811K}
\begin{tabular}{lrrrr}
\toprule
family & Del AUC (mean) & Del AUC (std) & Ins AUC (mean) & Ins AUC (std) \\
\midrule
densenet121 & 0.554 & 0.273 & 0.518 & 0.264 \\
resnet18\_cbam & 0.499 & 0.260 & 0.543 & 0.220 \\
swin\_tiny & 0.452 & 0.231 & 0.456 & 0.213 \\
vit\_tiny & 0.216 & 0.213 & 0.711 & 0.316 \\
\bottomrule
\end{tabular}
\end{table}

The family ordering is preserved and even slightly amplified (ViT Del 0.216 vs CNN 0.50--0.55). The faithfulness gap cannot be attributed to classification-accuracy differences.

\phantomsection
\subsection*{S12. Ablation: top-k sensitivity}
\addcontentsline{toc}{subsection}{S12. Ablation: top-k sensitivity}

To verify that the family ordering is not an artifact of the 10 \% perturbation threshold, we measure the confidence drop when removing only the top 5 \%, 10 \%, or 20 \% of heatmap-ranked pixels.

\begin{table}[htbp]
\centering
\small
\caption{\textbf{Supplementary Table S12.} Mean confidence drop ($\Delta$p) after removing top-k\% pixels, WM-811K}
\begin{tabular}{lrrr}
\toprule
family & top-5\% & top-10\% & top-20\% \\
\midrule
densenet121 & 0.106 & 0.142 & 0.238 \\
resnet18\_cbam & 0.029 & 0.046 & 0.136 \\
swin\_tiny & 0.034 & 0.086 & 0.210 \\
vit\_tiny & 0.362 & 0.492 & 0.589 \\
\bottomrule
\end{tabular}
\end{table}

ViT-Tiny Rollout's top pixels are prediction-critical at every granularity ($\Delta$p = 0.36 at 5 \%, 0.49 at 10 \%, 0.59 at 20 \%), while CNN Grad-CAM pixels produce minimal confidence change (ResNet $\Delta$p = 0.03--0.14). The family ordering is stable across all three thresholds, indicating that the result is not sensitive to the perturbation granularity.

\phantomsection
\subsection*{S13. Swin-Tiny supplementary details}
\addcontentsline{toc}{subsection}{S13. Swin-Tiny supplementary details}

\begin{center}
\small \textbf{Supplementary Table S13.} Swin-Tiny configuration and results summary
\end{center}

\begin{tabular}{lcc}
\toprule
Property & WM-811K (from scratch) & MVTec AD (pretrained) \\
\midrule
Parameters & 27,504,723 & 27,520,892 \\
Input size & 64$\times$64, 1-channel & 256$\times$256, 3-channel RGB \\
Window size & 4 & 7 \\
Pretrained & No & ImageNet \\
Seeds & 7, 123, 456 & 42 \\
Epochs (mean) & 88 & 57 \\
Balanced Accuracy & 82.2 $\pm$ 1.5\% & 93.7\% \\
F1 Macro & 0.792 $\pm$ 0.004 & 0.947 \\
Grad-CAM target & \texttt{layers[-1]} (final stage) & \texttt{layers[-1]} (final stage) \\
Native Del AUC & 0.432 $\pm$ 0.017 & 0.509 \\
Native Ins AUC & 0.480 $\pm$ 0.028 & 0.909 \\
Stability & 0.893 $\pm$ 0.006 & --- \\
RISE Del AUC & 0.096 & 0.709 \\
\bottomrule
\end{tabular}

\textbf{Architecture.} Swin-Tiny [23] uses shifted-window self-attention with 4 hierarchical stages. On WM-811K (64$\times$64, 1-channel), window size is set to 4 (yielding 4$\times$4 token windows at the first stage). On MVTec AD (256$\times$256, 3-channel RGB), the default window size of 7 is used. The model has 27.5M parameters --- larger than the other families (ResNet 11.2M, DenseNet 7.0M, ViT 5.4M) --- which partially explains its slower convergence from scratch.

\textbf{Seed selection (WM-811K).} Seed 42 produced a degenerate initialisation that failed to converge (BAcc 23.4\%, early-stopped at epoch 21). This is a known sensitivity of transformers trained from scratch on small datasets. Seed 7 was substituted, yielding BAcc 80.9\%. Seeds 123 and 456 converged normally (81.4\% and 84.2\% respectively). The three reported seeds (7, 123, 456) are consistent with each other (std 1.5\%), consistent with the seed-42 failure being an outlier rather than a systematic problem.

\textbf{Training dynamics.} From-scratch Swin-Tiny requires substantially more epochs than the CNNs (78--97 vs 28--36) and converges to a lower optimum (82\% vs 90\% BAcc). This is consistent with the general finding that transformers require more data or pretraining to match CNN performance on small datasets. On MVTec AD with ImageNet pretraining, Swin-Tiny matches the CNNs (93.7\% BAcc), indicating that the from-scratch gap is a data-efficiency issue rather than an architectural limitation.

\textbf{Grad-CAM target.} The final transformer stage (\texttt{model.backbone.layers[-1]}) is used as the Grad-CAM target layer. This stage produces a spatial feature map of shape (B, 2, 2, 768) for 64$\times$64 input, which is upsampled to input resolution. The \texttt{\_to\_4d()} utility handles Swin's (B, H, W, C) output format by permuting to (B, C, H, W) before standard Grad-CAM processing.

\phantomsection
\subsection*{S14. Code and data availability}
\addcontentsline{toc}{subsection}{S14. Code and data availability}

The complete source code, configuration files and evaluation scripts for both WM-811K and MVTec AD experiments are available at:

\textbf{Repository:} \href{https://github.com/JasperJia33/native-readout-xai}{github}    

The WM-811K dataset [18] is publicly available. MVTec AD [19] is available from the MVTec website under academic licence.

\end{document}